\documentclass[reprint,superscriptaddress,showpacs,amsmath,amssymb,aps]{revtex4-1}
\usepackage{floatrow}
\newfloatcommand{capbtabbox}{table}[][\FBwidth]
\usepackage{graphicx}
\usepackage{dcolumn}
\usepackage{bm}
\usepackage{hyperref}
\usepackage{subfigure}
\hypersetup{colorlinks=true, citecolor=blue, urlcolor=blue, linkcolor=blue}
\usepackage{amssymb}
\usepackage{color}
\usepackage{graphicx}
\usepackage{amsmath}
\usepackage{mathrsfs}
\usepackage{times}
\usepackage{subeqnarray}
\usepackage{cases}
\usepackage{bm}
\usepackage{diagbox}
\usepackage{booktabs}
\usepackage[table,xcdraw]{xcolor}
\usepackage{colortbl}
\usepackage{enumitem} 
\usepackage{tikz} 
\usepackage{epstopdf}
\definecolor{tabcolor}{rgb}{.105,.410,.113}
\definecolor{Pink}{RGB}{255,51,204}
\begin{document}

\title{Decoding fairness: a reinforcement learning perspective}
\author{Guozhong Zheng}
\affiliation{School of Physics and Information Technology, Shaanxi Normal University, Xi'an 710061, P. R. China}
\author{Jiqiang Zhang}
\affiliation{School of Physics, Ningxia University, Yinchuan 750021, P. R. China}
\author{Xin Ou}
\affiliation{School of Physics and Information Technology, Shaanxi Normal University, Xi'an 710061, P. R. China}
\author{Shengfeng Deng}
\affiliation{School of Physics and Information Technology, Shaanxi Normal University, Xi'an 710061, P. R. China}
\author{Li Chen}
\email[Email address: ]{chenl@snnu.edu.cn}
\affiliation{School of Physics and Information Technology, Shaanxi Normal University, Xi'an 710061, P. R. China}

\begin{abstract}
Behavioral experiments on the ultimatum game (UG) reveal that we humans prefer fair acts, which contradicts the prediction made in orthodox Economics. Existing explanations, however, are mostly attributed to exogenous factors within the imitation learning framework. Here, we adopt the reinforcement learning paradigm, where individuals make their moves aiming to maximize their accumulated rewards. Specifically, we apply Q-learning to UG, where each player is assigned two Q-tables to guide decisions for the roles of proposer and responder. In a two-player scenario, fairness emerges prominently when both experiences and future rewards are appreciated. In particular, the probability of successful deals increases with higher offers, which aligns with observations in behavioral experiments. 
Our mechanism analysis reveals that the system undergoes two phases, eventually stabilizing into fair or rational strategies.
These results are robust when the rotating role assignment is replaced by a random or fixed manner, or the scenario is extended to a latticed population.
Our findings thus conclude that the endogenous factor is sufficient to explain the emergence of fairness, exogenous factors are not needed.
\end{abstract}

\date{\today }
\maketitle
\section{Introduction}\label{sec:introduction}

Fairness is fundamental to the sustainable development of human beings. It fosters social cohesion and well-being by ensuring equitable distribution of resources and opportunities. However, the rising social inequality, manifested through unrest, violence, and tensions between socioeconomic groups, threatens these foundations~\cite{Piketty2014Capital,stiglitz2019people}. Therefore, it is crucial to decipher the mechanisms behind the emergence of fairness to achieve and sustain a decent level of fairness~\cite{Fehr2003TheNature}.

Evolutionary game theory provides a valuable framework for addressing this issue, and the ultimatum game (UG)~\cite{Guth1982An,Guth2014More} is the most widely adopted model for understanding fairness. In this game, two players -- one proposer and one responder -- decide how to divide a fixed amount of money. The proposer makes an offer, and the responder can either accept or reject it. If accepted, the money is distributed as proposed; if rejected, neither player receives anything. With the \emph{Homo economicus}~\cite{Simon1957Models,Samuelson2005Economics} assumption made in orthodox Economics, individuals are fully rational and self-interested; the responder will accept any nonzero offers since ``something is better than nothing", and the proposer will then propose the least possible offer to maximize the remaining proportion for its own. As such, an extremely unfair division is expected.
However, abundant behavioral experiments with UG show that most offers fall between $40\%$ and $50\%$ of the total, and offers below $20\%$ are often rejected~\cite{Guth1982An, Thaler1988Anomalies,Bolton1995Anonymity, Roth1995The, Guth2014More}. This indicates that humans strongly prefer fair acts, challenging the prediction made in orthodox Economics. 

To address this discrepancy, many theoretical studies have explored the mechanisms behind the emergence of fairness~\cite{Debove2016Models}. 
Research has revealed that the underlying structure of the population~\cite{Page2000The,Kuperman2008The,Sinatra2009The}, role assignment~\cite{Iranzo2011The,Deng2021Effects}, and noise~\cite{Gale1995Learning} significantly contribute to the emergence of fairness. 
Human factors like reputation~\cite{Nowak2000Fairness,Chiang2008Apath,ANDRE20111Social,ANDRE20111Social,Zhang2023Reputation}, spite~\cite{Forber2014The}, and empathy~\cite{Sanchez2005Altruism,Page2002Empathy,Attila2012Defense} are also found to be crucial for the evolution of fairness.
Interestingly, the presence of a few ``good Samaritans" can boost fairness remarkably via a discontinuous phase transition~\cite{zheng2022probabilistic} and may serve as a control strategy~\cite{Zheng2023Pinning}.
Notice that these studies typically employ the imitation learning framework~\cite{Roca2009Evolutionary,Szabo1998Evolutionary}, where individuals tend to imitate the strategies of neighbors who have higher payoffs. Imitation can be viewed as a simple form of social learning~\cite{Bandura1977social}, where people learn from others in socioeconomic activities by observations or instructions.

In contrast, reinforcement learning (RL)~\cite{Sutton2018reinforcement} offers a fundamentally distinct paradigm where individuals optimize their strategies through interaction with their environment. Unlike social learning, where decisions are made by observing and imitating peers, players with RL resort to self-reflection by evaluating their actions within different surroundings. This allows them to evolve unique and potentially heterogeneous strategies to better adapt to their surroundings rather than following uniform decision-making rules assumed in social learning. More importantly, individuals with RL aim to maximize the accumulated payoffs instead of immediate rewards as in social learning. Growing evidence shows that RL has distinct working areas and physiological processes in the brain~\cite{Lee2012neural, Rangel2008framework, Olsson2020neural}, manifesting itself as a fundamentally different learning paradigm compared to social learning.
Only recently, RL started to be utilized to study emergent behaviors such as cooperation~\cite{Tanabe2012Evolution,Ezaki2016Reinforcement,Horita2017Reinforcement,Ding2019Q,Fan2022Incorporating,Zhang2020Oscillatory,Wang2022Levy,Wang2023Synergistic,He2022migration,Ding2023emergence,Geng2022Reinforcement,Yang2024Interaction,Shi2022analysis,Zhang2024emergence,zheng2024evolution}, trust~\cite{Zheng2024Decoding}, resource allocation~\cite{Andrecut2001q, Zhang2019reinforcement, zheng2023optimal}, collective motion~\cite{Incera2020Development,Wang2023Modeling}, and other collective behaviors for humans~\cite{Tomov2021multi,Shi2022analysis}. 
For instance, Ding et al.~\cite{Ding2023emergence} revealed that individuals with historical experience and expected payoffs learn a ``win-stay, lose-shift" strategy to sustain decent cooperation in a two-agent scenario. Zheng et al.~\cite{Zheng2024Decoding} found that trust and trustworthiness arise when individuals appreciate past experiences and future returns. 
Given its great potential for deciphering many puzzles of human behaviors, it's intriguing to ask: \emph{Does the self-reflective RL learning paradigm provide a fresh perspective on the emergence of fairness, specifically whether the fair acts of people can be explained by endogenous motivation in humans?} 
This fundamentally differs from the current explanations attributed to the exogenous factors in the social learning paradigm.

In this study, we investigate the evolution of fairness within the RL paradigm. Specifically, we apply the Q-learning algorithm~\cite{Watkins1989learning,Watkins1992Q} to study the UG, where each individual is associated with two Q-tables: one for the role of a proposer and the other for the responder role. In a two-player scenario, fairness emerges significantly when individuals appreciate both past experiences and future rewards. Particularly, we find that the dependence of a successful deal on the offer is consistent with the observations made in behavioral experiments. 
Our mechanism analysis reveals a two-phase transition where only strategies of successful deals survive, and ultimately, their preferences evolve mainly to fair strategy together with a small amount of rational type.

This paper is organized as follows: Sec.~\ref{sec:model} introduces the Q-learning framework for the UG. Sec.~\ref{sec:results} presents the results for the two-player scenario, while Sec.~\ref{sec:analysis} provides a detailed mechanistic analysis. Sec.~\ref{sec:failure} discusses the failure of emergence. Sec.~\ref{sec:lattice} presents the robustness check and the extension to a latticed population. Finally, Sec.~\ref{sec:discussion} concludes with a summary and discussion of our findings.

\section{Model}\label{sec:model}

Let's consider a two-player UG scenario. As a rule of thumb, the total amount of money to be divided in each round is set to be 1. The strategy of each player $i$ is defined by $S_i=(p_i, q_i)$ with $p, q \in [0,1]$, where $p_i$ is the offer when acting as a proposer and $q_i$ is the acceptance threshold, the minimal offer expected when acting as a responder. 
Now, suppose two players, $i$ and $j$ are engaged in this game, the expected payoff for player $i$ is then:
\begin{equation}
\pi_p=\left\{\begin{array}{cl}
1-p_i, & \text { if } p_i \geqslant q_j, \\
0, & \text { otherwise},
\end{array}\right. \quad
\pi_r=\left\{\begin{array}{cl}
p_j, & \text { if } q_i \leqslant p_j, \\
0, & \text { otherwise}.
\end{array}\right.
\label{eq:payoff_UG}
\end{equation}
Here $\pi_{p,r}$ are the payoffs for individual $i$ and $j$ when they act as proposer and responder, respectively.
In the practice of many behavioral experiments~\cite{Lauren2024The,moretti2010disgust,wang2024rejecting,wang2024self}, three fixed offers are provided for proposers:  $p_l\!=\!l\!<\!0.5$, $p_m\!=\!m\!=\!0.5$, and $p_h\!=\!h\!>\!0.5$, corresponding to mean, fair, and overgenerous offers, respectively. Likewise, three typical acceptance thresholds are available for responders, and the values are the same as offers: $q_l=l$, $q_m=m$, and $q_h=h$. As discussed above, the rational strategy ($p_l$, $q_l$) is always preferred for \emph{Homo economicus}, which is an unfair outcome.

In our study, each player adopts the Q-learning algorithm to make decisions. Specifically, their decision-making is guided by two Q-tables, one for the proposer role and one for the responder role. The Q-table is a two-dimensional table expanded by states and actions shown in Table~\ref{tab:Qtable_UG}. 
The state set $\mathbb{S} = \{s_1, ..., s_9\}$, also the strategy set, consists of nine states by combination involving both roles' actions in the previous round. 
For example, the state $s_2 = (p_l, q_m)$ denotes that the proposer offers $l$ against the responder with the acceptance threshold $m$.
The action set is $\mathbb{A}_p = \{p_l, p_m, p_h\}$ for proposers, and $\mathbb{A}_r = \{q_l, q_m, q_h\}$ for responders.
The items $Q_{s,a}$ in the table are called the action-value function, estimating the value of the action $a$ within the given state $s$, which can be taken as a measure of action preference. The larger the value of $Q_{s,a}$, the stronger preference in action $a$ for the player within state $s$.

\begin{table}[t]
\centering
\scalebox{0.95}{
\begin{tabular}{c|ccc}
\toprule[0.5pt]
\hline
\diagbox{State}{Action}& $l$ ($a_{1}$) & $m$ ($a_{2}$) & $h$ ($a_{3}$) \\
\midrule [0.5pt]
\hline
$(p_l,q_l)$ ($s_{1}$) & $Q_{s_{1},a_{1}}$ & $Q_{s_{1},a_{2}}$ & $Q_{s_{1},a_{3}}$ \\
$(p_l,q_m)$ ($s_{2}$) & $Q_{s_{2},a_{1}}$ & $Q_{s_{2},a_{2}}$ & $Q_{s_{2},a_{3}}$ \\
$(p_l,q_h)$ ($s_{3}$) & $Q_{s_{3},a_{1}}$ & $Q_{s_{3},a_{2}}$ & $Q_{s_{3},a_{3}}$\\
$(p_m,q_l)$ ($s_{4}$) & $Q_{s_{4},a_{1}}$ & $Q_{s_{4},a_{2}}$ & $Q_{s_{4},a_{3}}$ \\
$(p_m,q_m)$ ($s_{5}$) & $Q_{s_{5},a_{1}}$ & $Q_{s_{5},a_{2}}$ & $Q_{s_{5},a_{3}}$ \\
$(p_m,q_h)$ ($s_{6}$) & $Q_{s_{6},a_{1}}$ & $Q_{s_{6},a_{2}}$ & $Q_{s_{6},a_{3}}$\\
$(p_h,q_l)$ ($s_{7}$) & $Q_{s_{7},a_{1}}$ & $Q_{s_{7},a_{2}}$ & $Q_{s_{7},a_{3}}$ \\
$(p_h,q_m)$ ($s_{8}$) & $Q_{s_{8},a_{1}}$ & $Q_{s_{8},a_{2}}$ & $Q_{s_{8},a_{3}}$ \\
$(p_h,q_h)$ ($s_{9}$) & $Q_{s_{9},a_{1}}$ & $Q_{s_{9},a_{2}}$ & $Q_{s_{9},a_{3}}$\\
\hline
\bottomrule[0.5pt]
\end{tabular}
}
\caption{\textbf{Q-table structure in the 2-player scenario.} The state is defined by the action combination of the proposer and responder in the previous round, and three options are available for taking action. The two Q-tables for the proposer and responder when the action set is conveniently replaced with $\mathbb{A}_{p}$ and $\mathbb{A}_{r}$, respectively.}
\label{tab:Qtable_UG}
\end{table}

\begin{figure*}[!tbp]
\centering
\includegraphics[width=0.8\linewidth]{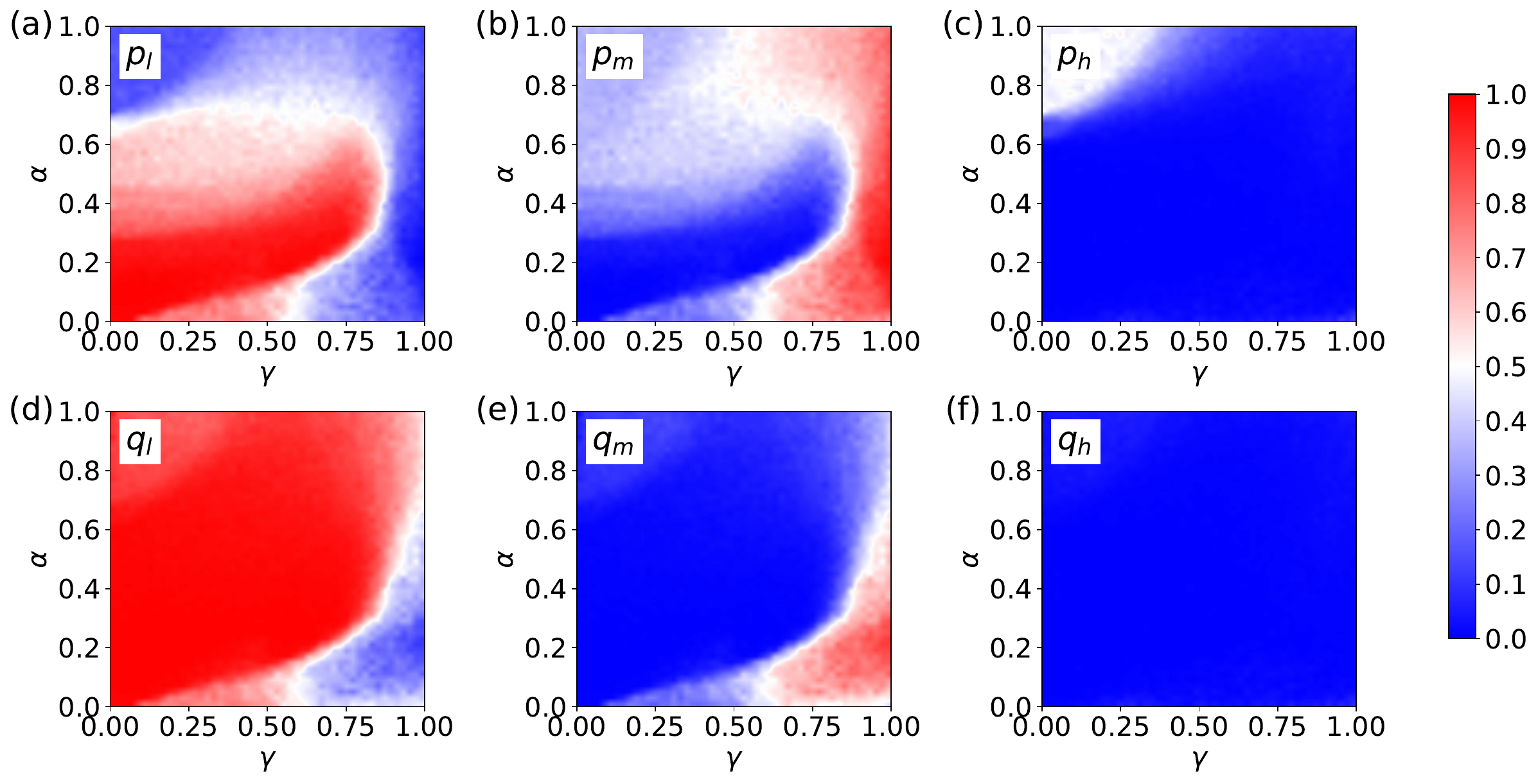}
\caption{\textbf{Emergence of fairness.} 
The color-coded stationary fractions of different options in the learning parameter domain ($\gamma,\alpha$), both ranging from 0 to 1. Panels (a-c) represent the fractions $f_{p_{l,m,h}}$ for the role of the proposer, and panels (d-f) show $f_{q_{l,m,h}}$ for the responder, respectively. Large fractions of $f_{p_m}$ and $f_{q_m}$ are both seen in the corner of large $\gamma$ and small $\alpha$, showing that fairness emerges when individuals value historical experience and have a long-term vision. Otherwise, either forgetful (with a large $\alpha$) or shortsighted (with a small $\gamma$) individuals typically choose the rational strategy, resulting in fractions of $f_{p_l}$ and $f_{q_l}$. Each data is averaged over 100 realizations. Other parameters: $\epsilon=0.01$, $l=0.3$, and $h=0.8$.
}
\label{fig:phasediagram_0.30.50.8}
\end{figure*}

Without loss of generality, the two players are initially assigned random actions from $\mathbb{A}_{p,r}$, and their roles alternate from round to round. In round $t$, both players independently select an action $a$ from their action sets with a probability of $\epsilon$, engaging in a random exploration. Otherwise, they choose the action $a$ with the highest Q-value within the current state $s$ from the associated Q-table for their roles. After making decisions, each player learns from this experience by updating the Q-value of the action-state pair just used with the following Bellman equation:
\begin{equation}
\begin{aligned}
Q_{s, a}(t\!+\!1) & =Q_{s, a}(t)\!+\!\alpha\left(\pi(t)\!+\!\gamma \max _{a^{\prime}} Q_{s^{\prime}, a^{\prime}}(t)\!-\!Q_{s, a}(t)\right) \\
& =(1\!-\!\alpha) Q_{s, a}(t)\!+\!\alpha\left(\pi(t)\!+\!\gamma \max _{a^{\prime}} Q_{s^{\prime}, a^{\prime}}(t)\right),
\end{aligned}
\label{eq:Qlearning}
\end{equation}
where $s$ and $a$ denote the current state and action just taken by the focal individual, $s^{\prime}$ represent the new state for the round $t+1$. $\pi(t)=\pi_p$ or $\pi_r$ up to the role of the focal player and is computed according to Eq.~(\ref{eq:payoff_UG}). The parameter $\alpha\in(0,1]$ is the learning rate, which determines the contribution to Q-value from the current step. A larger value of $\alpha$ indicates that players are more ``forgetful," as the old Q-values are erased more rapidly; individuals with a small $\alpha$ better reserve the experience instead. 
$\gamma\in[0,1)$ is the discount factor, which captures the weight of future rewards, where $\max _{a^{\prime}} Q_{s^{\prime}, a^{\prime}}(t)$ denotes the expected maximum value in the next round. Individuals with a larger value of $\gamma$ can be interpreted as having a long vision; otherwise, they prioritize the immediate rewards and thus can be considered shortsighted.  This completes the learning process and the evolution at the round $t$. For clarity, the evolution protocol is summarized in Appendix~\ref{sec:appendixA}.

In this study, we compute the fractions of different offers and acceptance thresholds at the round $\tau$, defined as:
\begin{equation}
\begin{aligned}
f_p(\tau):=\frac{1}{N} \sum_{i=1}^N \mathbb{I}\left(a_i(\tau)=p\right), \quad p \in\mathbb{A}_p,\\
f_q(\tau):=\frac{1}{N} \sum_{i=1}^N \mathbb{I}\left(a_i(\tau)=q\right), \quad q \in\mathbb{A}_r.    
\end{aligned}
\end{equation}
$\mathbb{I}(\mathbb{X})$ is the indicator function that takes the value 1 if the associated condition $\mathbb{X}$ is satisfied and 0 otherwise. $N$ is the population size and $N=2$ in our case.
For reliable results, we also conduct the ensemble average given by
\begin{equation}
\begin{aligned}
\bar{f}_p(\tau) := \frac{1}{M} \sum_{k=1}^M f_p^k(\tau), \quad p \in\mathbb{A}_p, \\
\bar{f}_q(\tau) := \frac{1}{M} \sum_{k=1}^M f_q^k(\tau), \quad q \in\mathbb{A}_r.
\end{aligned}
\end{equation}
Here, $f_{p,q}^k (\tau)$ are the observed values of \( f_{p,q} \) at round $\tau$ during the \( k \)-th realization, and \( \bar{f}_{p,q} \) represent their ensemble averages over \( M \) independent realizations. The fairness level can be assessed by primarily monitoring \( \bar{f}_{p_m} \) and \( \bar{f}_{q_m} \), which represent the average fraction adopting $p_m$ and $q_m$, respectively.
In our practice, the data is time-averaged over 1000 rounds after a transient of $1 \times 10^8$ steps for each realization, and we remove the bar and simply use $f_{p,q}$ to denote the ensemble results for notation clarity.

\section{Results}\label{sec:results}
\begin{figure*}[!tbp]
\centering
\includegraphics[width=0.8\linewidth]{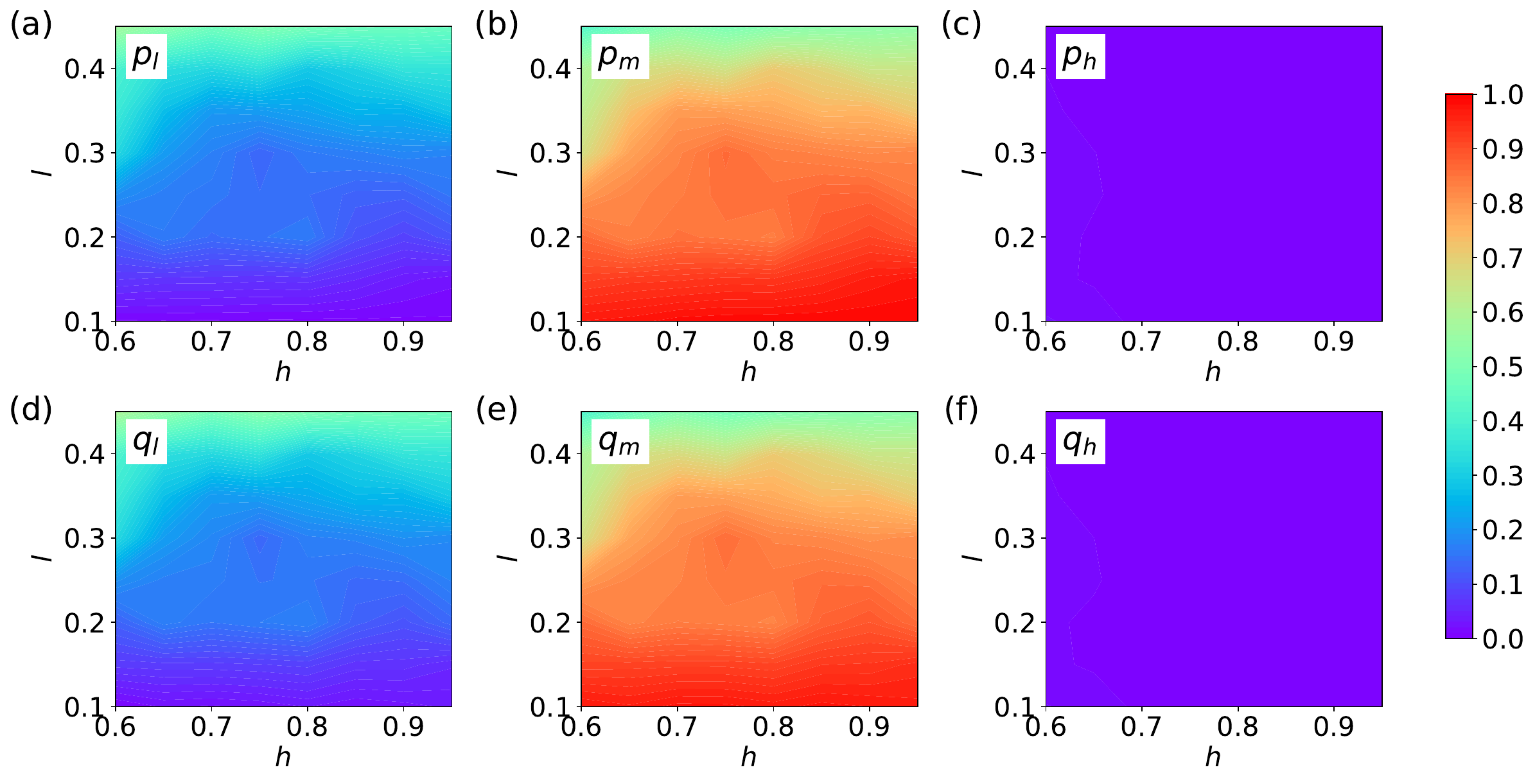}
\caption{\textbf{Dependence on the game parameters.} The color-coded dependence of different options on the two control parameters $l$ and $h$. Panels (a-c) are the average fractions for three offers $f_{p_{l,m,h}}$ for the proposer, and panels (d-f) are three acceptance thresholds $f_{q_{l,m,h}}$ for the responder, respectively. Each data is averaged 100 realizations. 
Panels (c) and (f) show that both $f_{p_h}$ and $f_{q_h}$ remain below 0.05, while the option fractions are predominantly concentrated on the rational options $p_l,q_l$ and fair options $p_m,q_m$. As $l$ increases, the fractions of options $p_l$ and $q_l$ gradually rise.
Other parameters: $\epsilon=0.01$, $\alpha=0.1$, and $\gamma=0.9$.}
\label{fig:heatmap2}
\end{figure*}

We first report the emergence of fairness in the learning parameter domain, shown in Fig.~\ref{fig:phasediagram_0.30.50.8}. By fixing $l=0.3$ and $h=0.8$, the shown provides the densities of three preferences for proposers and responders in the upper and lower panels, respectively. We find that fairness emerges prominently in a physically meaningful region. As shown in Fig.~\ref{fig:phasediagram_0.30.50.8}(b,e), when the learning rate $\alpha$ is small, and the discount factor $\gamma$ is large, the fractions of $p_m$ and $q_m$ approach unity (red region). This means that when individuals appreciate historical experiences (a small $\alpha$) and have long-term visions (a large $\gamma$), they are more likely to propose and expect a fair offer.
Otherwise, people with either a large $\alpha$ (forgetful) or a small $\gamma$ (shortsighted) typically prefer the rational strategy $(p_l, q_l)$, shown in Fig.~\ref{fig:phasediagram_0.30.50.8}(a,d). Notice that the fraction of overgenerous offer $p_h$ and acceptance threshold $q_h$ are almost vanishing. Only when individuals are both forgetful and shortsighted may they propose such an offer.
These observations qualitatively hold for different game parameters, an example for $l=0.4$ and $h=0.6$ are provided in Appendix~\ref{sec:appendixB}. 

To systematically examine the impact of two game parameters, we present the dependence of the six fractions on $l$ and $h$ by fixing the two learning parameters at $\alpha=0.1$ and $\gamma=0.9$, a parameter combination within the red region in Fig.~\ref{fig:phasediagram_0.30.50.8}(b).
Fig.~\ref{fig:heatmap2}(c,f) show that the overgenerous offer is rarely proposed or expected, as both densities of $p_h$ and $q_h$ remain close to zero across the entire domain. Consequently, the value $h$ is supposed to have a marginal impact on the evolution. This is indeed what we observe in the other four subplots in Fig.~\ref{fig:heatmap2}, where the fractions are insensitive to the magnitude change of $h$. By contrast, substantial changes in the four fractions are seen as $l$ is varied. With a very small value of $l$, e.g. $l=0.1$ corresponding to a low offer and expected offer, fair options dominate as  $f_{p_m,q_m} \rightarrow 1$. As $l$ is increased, e.g. $l=0.25$,  $f_{p_m,q_m}$ are reduced to around $80\%$ while the fractions of rational options rise to around $20\%$. The continual increase in $l$ further reduces the fraction of fair options, and the fraction of rational strategy is augmented instead. At $l=0.45$, the fractions for fair and rational options now become roughly equal.

\begin{figure}[!htbp]
\centering
\includegraphics[width=0.9\linewidth]{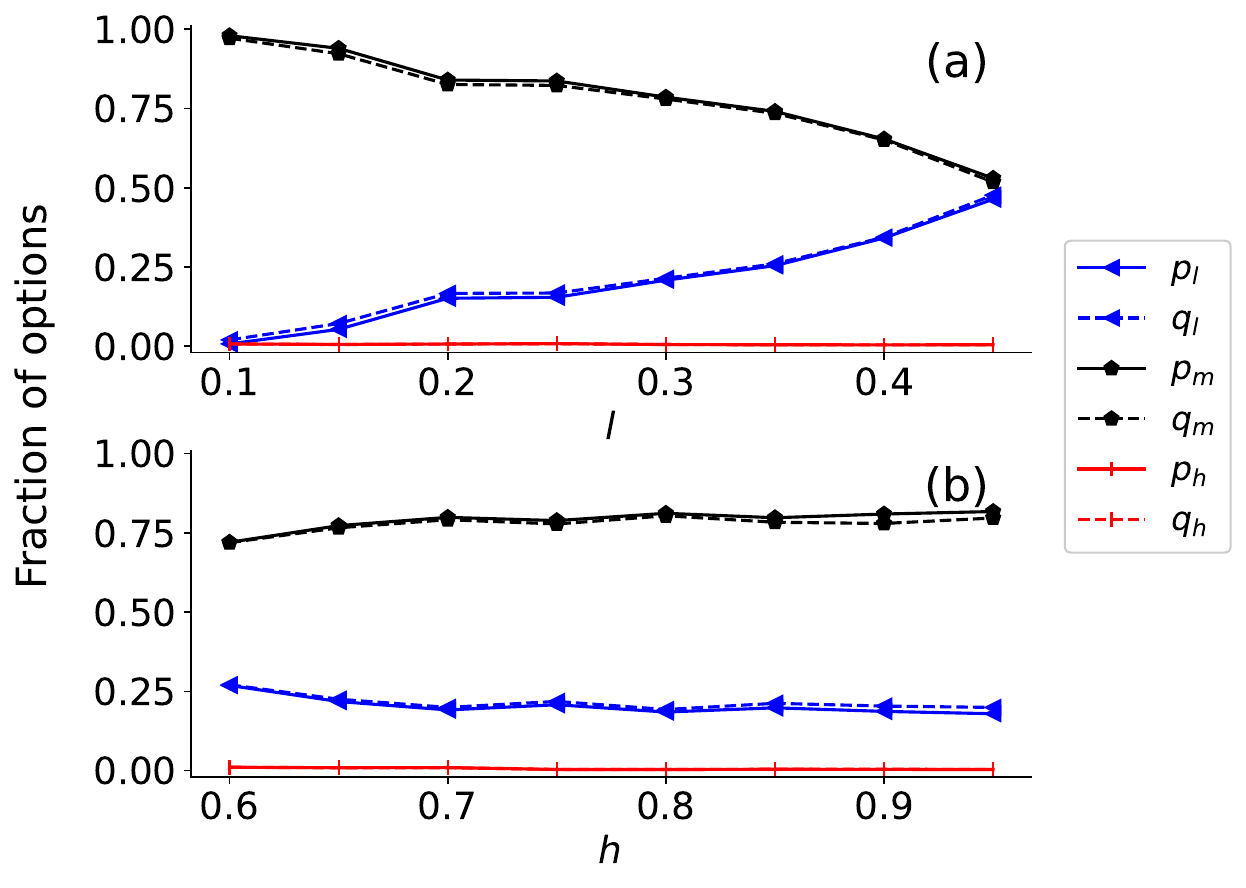}
\caption{\textbf{Distinct impacts of the two game parameters ($l$ and $h$) on the stationary fractions of options}. 
(a) The stationary fractions of six options versus $l$ by averaging over the row data of Fig.~\ref{fig:heatmap2}. 
(b) The stationary fractions of six options versus $h$ by averaging over the column data of Fig.~\ref{fig:heatmap2}.  
The two subplots show different dependencies of fairness on the $l$ and $h$. While the rational option fractions $f_{{p_l},{q_l}}$ rise as $l$ increases, the marginal impact of $h$ is observed. In all cases, the densities of overgenerous options are always vanishing.
}
 \label{fig:varylh}
\end{figure}

The distinct dependence of the two parameters in this game is well captured by doing row-average and column-average of Fig.~\ref{fig:heatmap2}.
Fig.~\ref{fig:varylh}(a) clearly shows that the declining trend for the fair options $f_{{p_m},{q_m}}$ as $l$ is increased, and this trend is reversed for the rational options $f_{{p_l},{q_l}}$.  These dependencies seem to become the opposite by varying $h$ shown in Fig.~\ref{fig:varylh}(b), but the trends are too weak to be claimed. Interestingly, Fig.~\ref{fig:varylh}(a) reveals that offers between $45\%$ and $50\%$ are more likely to be accepted, while offers below $20\%$ will be rejected with a probability of $80\%$. These results suggest a propensity for individuals to reject highly unfair offers, aligning with findings from previous behavioral experiments~\cite{Guth2014More}.

\begin{figure*}[!htbp]
\centering
\includegraphics[width=0.8\linewidth]{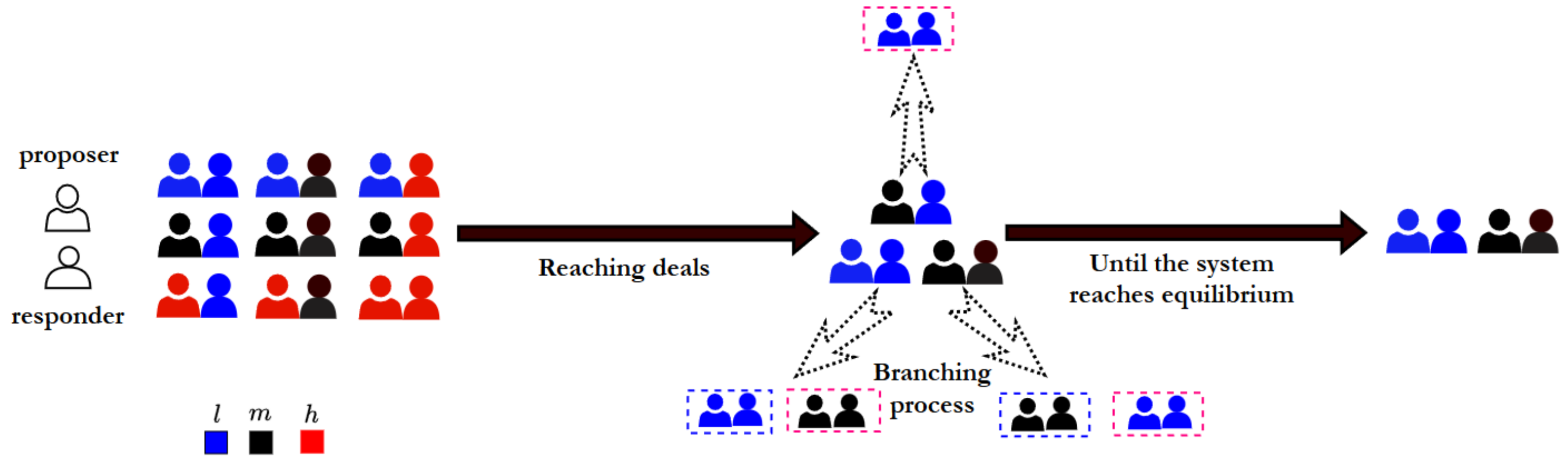}
\caption{\textbf{Schematic diagram of fairness emergence.} 
Two players start from all nine possible combinations due to the three options for either role, represented by blue, black, and red. The emergence of fairness experiences two phases.
In phase I, among nine combinations, the case with the failed deal cannot persist, only three of them $(p_m,q_m)$, $(p_l,q_l)$, and $(p_m,q_l)$ survive in this stage.  
In phase II, there are dominating transitions indicated by arrows among the remaining three strategies, reducing the fraction of $(p_m,q_l)$. Ultimately, only fair strategies $(p_m,q_m)$ and rational strategies $(p_l,q_l)$ survive. 
Red dashed boxes represent state transitions, while blue dashed boxes show the system tends to retain the current state.}
\label{fig:mechanism}
\end{figure*}

\section{Mechanism Analysis} \label{sec:analysis}
We now turn to deciphering the mechanism behind the fairness emergence.  
Roughly speaking, there are two phases behind the fairness emergence, as depicted in Fig.~\ref{fig:mechanism}.
 \begin{enumerate}[label=(\Roman*)]
\item Starting from random configurations, individuals revise their strategies to maximize the reward and improve the likelihood of successful deals. Those who propose and expect high offers put themselves in a disadvantageous position, and the strategy $(p_l,q_m)$ also fails to reach a deal. As a result, only three strategies $(p_m,q_m)$, $(p_l,q_l)$, and $(p_m,q_l)$ survive in this stage.
\item A dominating transition mode $(p_m,q_l)$ $\rightarrow$ $(p_l,q_l)$ $\rightarrow$$(p_m,q_m)$ is revealed, whereby the strategy $(p_m,q_l)$ vanishes. Ultimately,  the two players hold either the fair strategy $(p_m,q_m)$ or the rational strategy $(p_l,q_l)$.
\end{enumerate}

\subsection{Phase I -- Reaching deals}

\begin{figure*}[!htbp]
\centering
\includegraphics[width=0.85\linewidth]{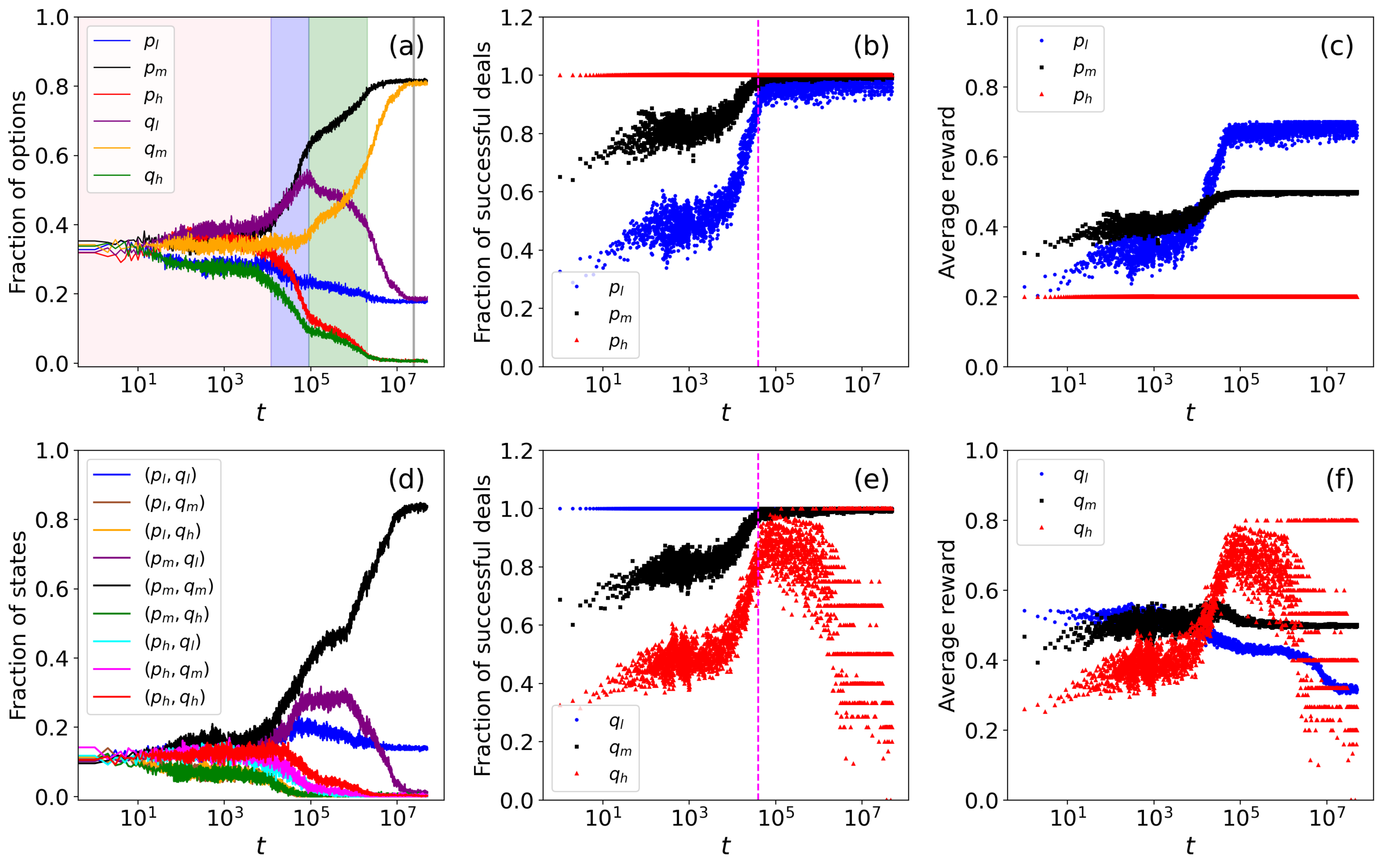}
\caption{\textbf{Typical time series in the 2-player scenario.}
(a) Time series for the fractions of six options.
(b, e) Time series for the fractions of successful deals for proposers and responders using the three options, respectively.
(c, f) Evolution of average payoffs for proposers and responders when choosing from the three options.
(d) Time series for the fractions of all nine states.
All data are averaged over 500 realizations. Other parameters: $\epsilon=0.01$, $\alpha=0.1$, $\gamma=0.9$, $l=0.3$, $h=0.8$.
}
\label{fig:TimeSeries}
\end{figure*}

To understand the evolution in Phase I, we present the typical time series for the case of $\alpha=0.1$, $\gamma=0.9$, and $(l, h)=(0.3, 0.8)$. Initially, the densities of all three options are evenly distributed for both the proposer and responder due to random initialization (around 0.33) [Fig.~\ref{fig:TimeSeries}(a)]. Over time, the fraction of $q_h$ first declines as its probability of successful deals is the lowest [Fig.~\ref{fig:TimeSeries}(e)], leading to the lowest average payoff [Fig.~\ref{fig:TimeSeries}(f)]. As a result, the option of $q_h$ is gradually abandoned by responders.

The fraction of $p_h$ is also found to decrease in Fig.~\ref{fig:TimeSeries}(a). Although $p_h$ ensures a successful deal [Fig.~\ref{fig:TimeSeries}(b)], it brings less favorable payoffs for proposers [Fig.~\ref{fig:TimeSeries}(c)]. As $q_h$ diminishes, proposers with a lower offer ($p_l$ or $p_m$) receive a higher reward and are better off. That is why the fraction of $p_h$ declines. The decline of $f_{p_h, q_h}$ also explains the vanishing fractions of the associated states in Fig.~\ref{fig:TimeSeries}(d). Notice that, the strategy $s_2=(p_l, q_m)$ also goes extinct due to the consistent deal failure.

At the end of Phase I ($\sim6\times10^4$, as indicated by the pink dashed line in Fig.~\ref{fig:TimeSeries}), the probability of successful deals approaches 1. The remaining combinations $s_1=(p_l, q_l)$, $s_4=(p_m, q_l)$, and $s_5=(p_m, q_m)$ [Fig.~\ref{fig:TimeSeries}(d)] are all able to reach successful deals.

\subsection{Phase II --  Branching process}

To monitor the evolution of the three remaining strategies (i.e. $s_{1,4,5}$) in Phase II, we track individuals' preferences within these states. This is captured by the fraction of options corresponding to $Q_{max}$ in the Q-tables, which provides the details for their competition and transition.

\begin{figure*}[!htbp]
\centering
\includegraphics[width=0.8\linewidth]{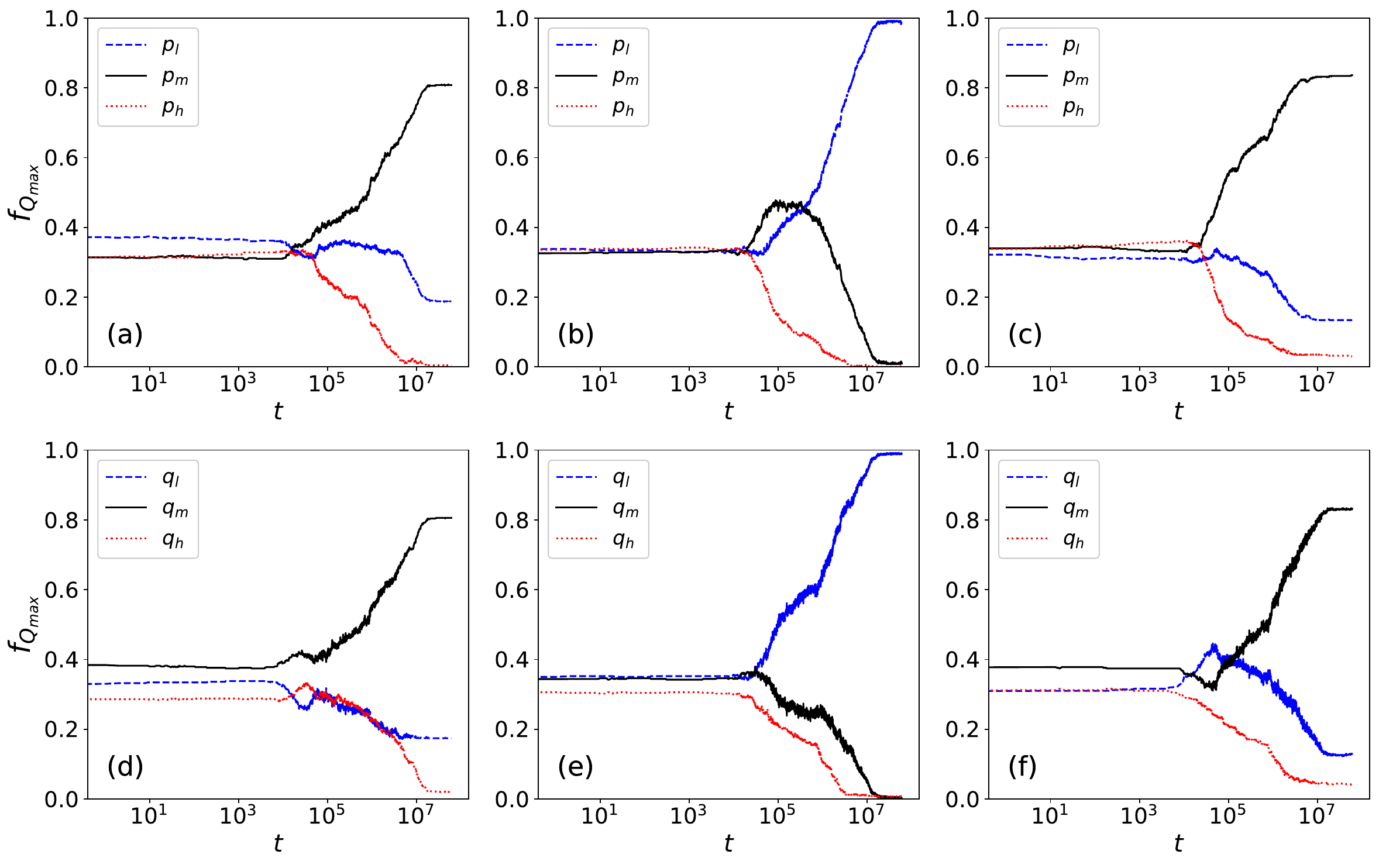}
\caption{\textbf{Evolution of preference.} 
The evolution of action preference captured by the maximal Q-value within each of the three dominating states $s_1=(p_l, q_l)$, $s_4=(p_m, q_l)$, and $s_5=(p_m, q_m)$. 
The preference evolution for proposers (a-c) and responders (d-f) within states $s_{1,4,5}$, respectively.
Each data is averaged over 500 realizations. Other parameters: $\epsilon=0.01$, $\alpha=0.1$, $\gamma=0.9$, $l=0.3$, and $h=0.8$.}
\label{fig:TimeSeries_Qmax}
\end{figure*}

As shown in Figs.~\ref{fig:TimeSeries_Qmax}(a,d), when individuals are within the state $s_1=(p_l,q_l)$, both the proposer and responder develop one of the two preferences ($m$ or $l$) and stick to them. Guided by the higher future reward, the responder tries $q_m$ to prompt the proposer to raise the offer. However, there are two branching outcomes. i) This choice of $q_m$ successfully forces the proposer to change the offer to $p_m$, and this fair strategy $(p_m,q_m)$ is stabilized as both players continuously get rewards. ii) The expected offer $q_m$ fails to meet, the combination $(p_l, q_m)$ leads to a failed deal, and the responder goes back to $q_l$, which yields continuous rewards within $(p_l,q_l)$ and also becomes stable. Ultimately, two states stabilize: turning to the fair strategy by $(p_l,q_l)\rightarrow (p_m,q_m)$ (two black lines), or keeping the status quo $(p_l,q_l)\rightarrow (p_l,q_l)$ (two blue lines).

By contrast, within the state $s_4=(p_m,q_l)$, the responder develops a dominating preference in $q_l$ as its fraction $f_{Q_{max}}\rightarrow 1$, shown in Fig.~\ref{fig:TimeSeries_Qmax}(e).  This is a bit counterintuitive as shifting from $q_l$ to $q_m$ avoids the reduction in offer by the proposer. This is true, but once the action of $q_m$ is chosen, the state is then changed.  Instead, by retaining the preference $q_l$, the responder obtains exactly the same immediate reward as the case of choosing $q_m$. What's different here is that the state remains the same, where the corresponding Q-value $Q_{s_4, q_l}$ is repeatedly reinforced and the preference in $q_l$ is further strengthened.  
In response, the proposer shifts the offer from $p_m$ to $p_l$, since the deal is still reached but with a higher reward [Fig.~\ref{fig:TimeSeries}(c)].
This preference transition of $(p_m,q_l)\rightarrow (p_l,q_l)$ is expected [Figs.~\ref{fig:TimeSeries_Qmax}(b,e)].

Finally, there is also a branching process observed within the state $s_5=(p_m,q_m)$ [Fig.~\ref{fig:TimeSeries}(c,f)]. i) By exploration, the responder lowers the expected offer to $q_l$, where the deal is still successful, and the obtained reward remains the same as in the case of $q_m$. The proposer then shifts the offer to $p_l$ as a response, and the combination $(p_l,q_l)$ is stabilized. ii) Staying in $(p_m,q_m)$ yielding continuous rewards, where the corresponding $Q_{s_5, p_m}$ and $Q_{s_5, q_m}$ are reinforced and the preference in fair options is stabilized. 
Therefore, $(p_m,q_m)\rightarrow (p_m,q_m)$ and $(p_m,q_m)\rightarrow (p_l,q_l)$ are expected in $s_5$ [Figs.~\ref{fig:TimeSeries_Qmax}(c,f)].
Notice that the branching processes in $s_{1,5}$ can only occur in the early stage of evolution; once the preferences are formed, the exploration event is hard to change them anymore. So, it's not a typical bistable state where the two states $s_{1,5}$ jump into each other; instead, it's a branching process, once one of the two states is reached, the system sticks to it. 

\begin{figure}[!t]
\centering
\includegraphics[width=1.0\linewidth]{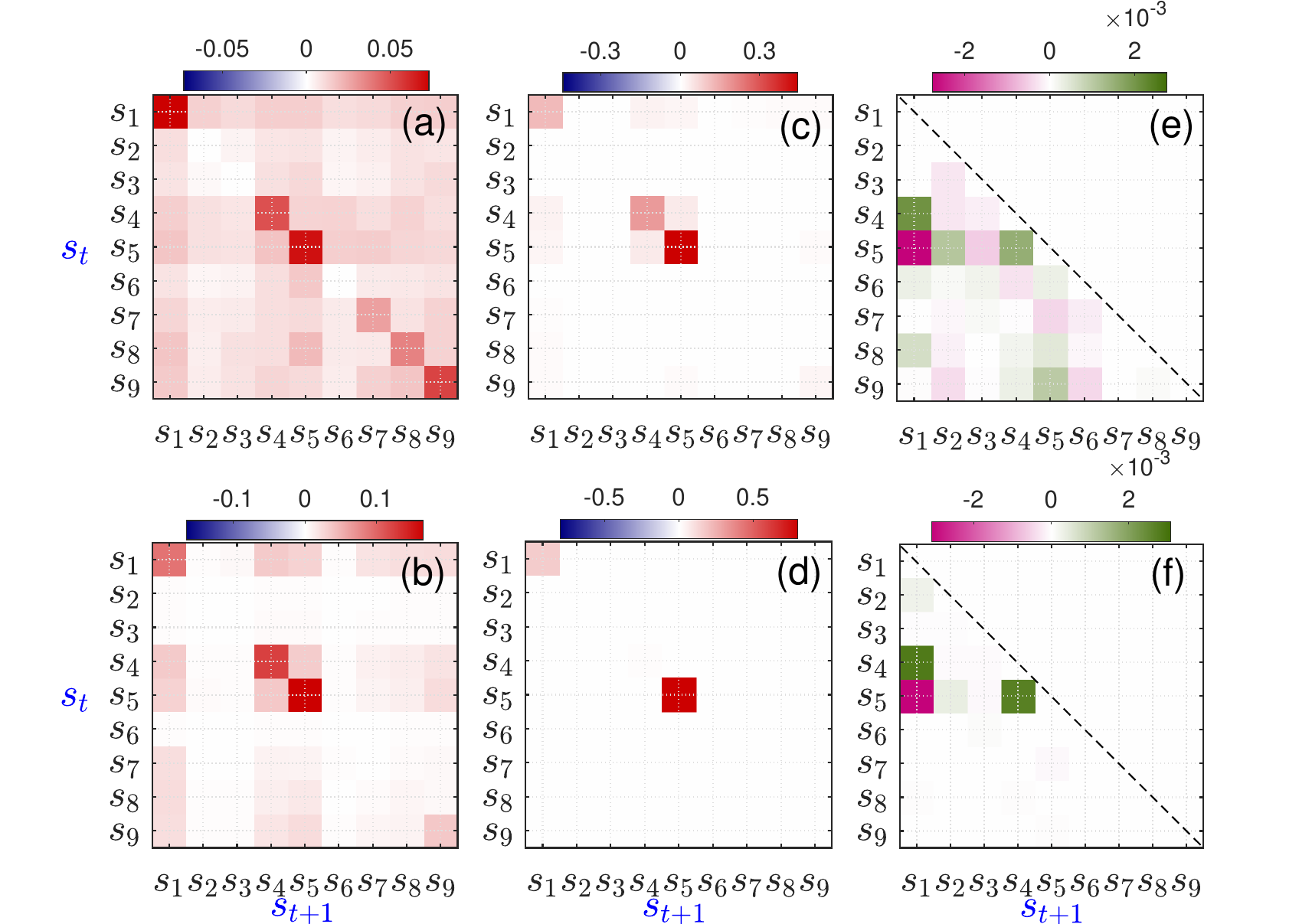}
\caption{\textbf{Joint probability matrices for two consecutive states.} The joint probability is defined as $P_{i,j}=\!P(s_{t}\!=\!s_i,s_{t+1}\!=\!s_j)$ for all 81 state transition paths for four typical stages, corresponding to the shaded regions highlighted in Fig.~\ref{fig:TimeSeries}: 
(a) $0\sim\!1.2\!\times10^4$, (b) $1.2\times10^4\!\sim\!9\times10^4$, (c) $9\times10^4\!\sim\!2\times10^6$, and (d) $2.3\times10^7\!\sim\!2.5\times10^7$. 
The states are defined in Tab.~\ref{tab:Qtable_UG}.
(e) and (f) show the probability difference defined by $\Delta P_{ij}\!=\!P_{i,j}\!-\!P_{j,i}$, respectively for (c) and (d). Only the data in the lower-left corner is shown.
Among the three states $s_{1,4,5}$ of our interest,  $s_4$ cannot stably persist, leaving the dominating fair strategy $s_5$ together with a small fraction of rational strategy $s_1$. 
Each data is averaged over 500 realizations. Other parameters: $\epsilon=0.01$, $\alpha=0.1$, $\gamma=0.9$, $l=0.3$, and $h=0.8$.}
\label{fig:State_transiton}
\end{figure}

To further validate the above analysis, we calculated the joint probabilities of two consecutive states defined as $P_{i,j}\!=\!P(s_{t}\!=\!s_i,s_{t+1}\!=\!s_j)$ for all $9\!\times\!9$ state combinations, as shown in Fig.~\ref{fig:State_transiton}. These probabilities capture state transitions over time and pin down frequent paths. In the early stage ($0<t\leq12000$, see Fig.~\ref{fig:State_transiton}(a)), almost all state transitions are observable, but some self-loops $s_1$$\rightarrow$$s_1$, $s_4$$\rightarrow$$s_4$, $s_5$$\rightarrow$$s_5$, $s_9$$\rightarrow$$s_9$, are more likely to occur, where the deals are all able to be reached.
As evolution progresses, three transitions $s_1$$\rightarrow$$s_1$, $s_4$$\rightarrow$$s_4$, $s_5$$\rightarrow$$s_5$ become dominant [Fig.~\ref{fig:State_transiton}(b)]. These findings align with our analysis in Phase I. Specifically, the pink- and purple-shaded regions in Fig.~\ref{fig:TimeSeries}(a) correspond to the strategy adjustment in the initial evolution stage before reaching deals.

Fig.~\ref{fig:State_transiton}(c) shows the evolution in Phase II, corresponding to the green-shaded region in Fig.~\ref{fig:TimeSeries}(a), where only transitions among $s_{1,4,5}$ are observable.  Apart from the above dominating transitions, we focus on $s_1$$\leftrightarrow$$s_4$, $s_1$$\leftrightarrow$$s_5$, and $s_4$$\leftrightarrow$$s_5$.                     
To obtain the net flow for these transitions, we compute the probability difference $\Delta P_{ij}\!=\!P_{i,j}\!-\!P_{j,i}$ for each pair of two states, where $\Delta P_{ij}>0$ means the net flow from $s_i$ to $s_j$ and vice versa.
Fig.~\ref{fig:State_transiton}(e) shows the net flows for Fig.~\ref{fig:State_transiton}(c), where their magnitudes are much lower. Three of them are frequent: $s_4$$\rightarrow$$s_1$, $s_1$$\rightarrow$$s_5$, and $s_5$$\rightarrow$$s_4$, though the latter is the weakest among them. This confirms the transitions: $(p_m,q_l)$$\rightarrow$$(p_l,q_l)$$\rightarrow$$(p_m,q_m)$ in Phase II. 
Eventually, the transition $s_4$$\rightarrow$$s_4$ disappears, with only $s_5$$\rightarrow$$s_5$ and $s_1$$\rightarrow$$s_1$ are left, as shown in Fig.~\ref{fig:State_transiton}(d). This corresponds to the gray-shaded region in Fig.~\ref{fig:TimeSeries}(a), where the fractions of all six options are stabilized. The corresponding net flow plot [Fig.~\ref{fig:State_transiton}(f)], however, shows some intricacies behind it. 
By exploration, the fair strategy may have $s_5$$\rightarrow$$s_4$, which is then taken advantage of by the proposer and leads to $s_4$$\rightarrow$$s_1$, and finally returns to the fair strategy through $s_1$$\rightarrow$$s_5$.

The above analysis applies to cases of different $l$ and $h$. As $l$ increases, the competitive advantage of the strategy $(p_l, q_l)$ grows, leading to an increase in its fraction, as observed in Fig.~\ref{fig:varylh}(a). Since the options of $q_h$ and $p_h$ are eliminated in the early stages due to failed deals, the resulting fractions are largely unaffected by $h$.
However, once the value of $h$ becomes close to $m=0.5$, the elimination of the $p_h$ option slows down. If that occurs, the presence of these three strategies $(p_h, q_{l,m,h})$ prompts the responder to choose $q_l$, which then facilitates the transition to $s_1$. That explains the slight increase in the fractions of $p_l$ and $q_l$ as $h$ approaches 0.6 in Fig.~\ref{fig:varylh}(b).

\begin{figure}[b]
\centering
\includegraphics[width=0.9\linewidth]{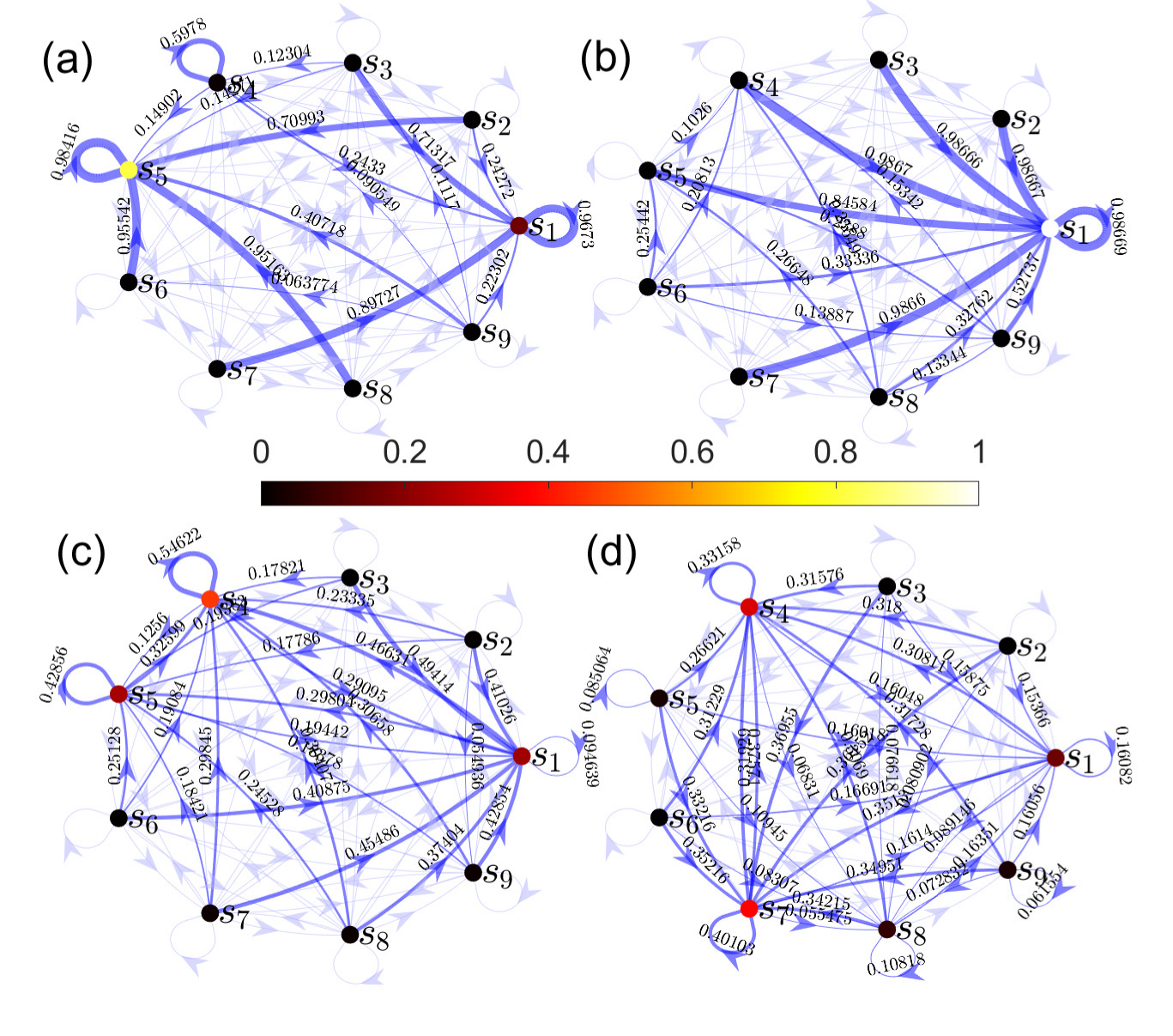}
\caption{\textbf{Four typical state transition networks.} 
The learning parameter combinations are for: (a) $(\alpha,\gamma)=(0.1,0.9)$, (b) $(\alpha,\gamma)=(0.1,0.1)$, (c) $(\alpha,\gamma)=(0.9,0.9)$, (d) $(\alpha,\gamma)=(0.9,0.1)$. 
In these networks, nodes represent states, with state probabilities $p(s)$ is color-coded.
The directed edges correspond to state transitions, with the magnitude of transition probabilities $p(s\rightarrow s')=p(s'|s)$ encoded in the line color and width. For clarity, transition probabilities below 0.05 are not shown. Note that $s_1=(p_l,q_l)$ and $s_5=(p_m,q_m)$ are, respectively, the rational and fair strategies of our particular interest.
Each data is averaged over 500 realizations, and for $2\times10^6$ time average after a transient of $2.3\times10^7$ steps.
Other parameters: $\epsilon=0.1$, $l=0.3$, and $h=0.8$.}
\label{fig:TransitionNetwork}
\end{figure}

\begin{figure}[!tbp]
\centering
\includegraphics[width=0.8\linewidth]{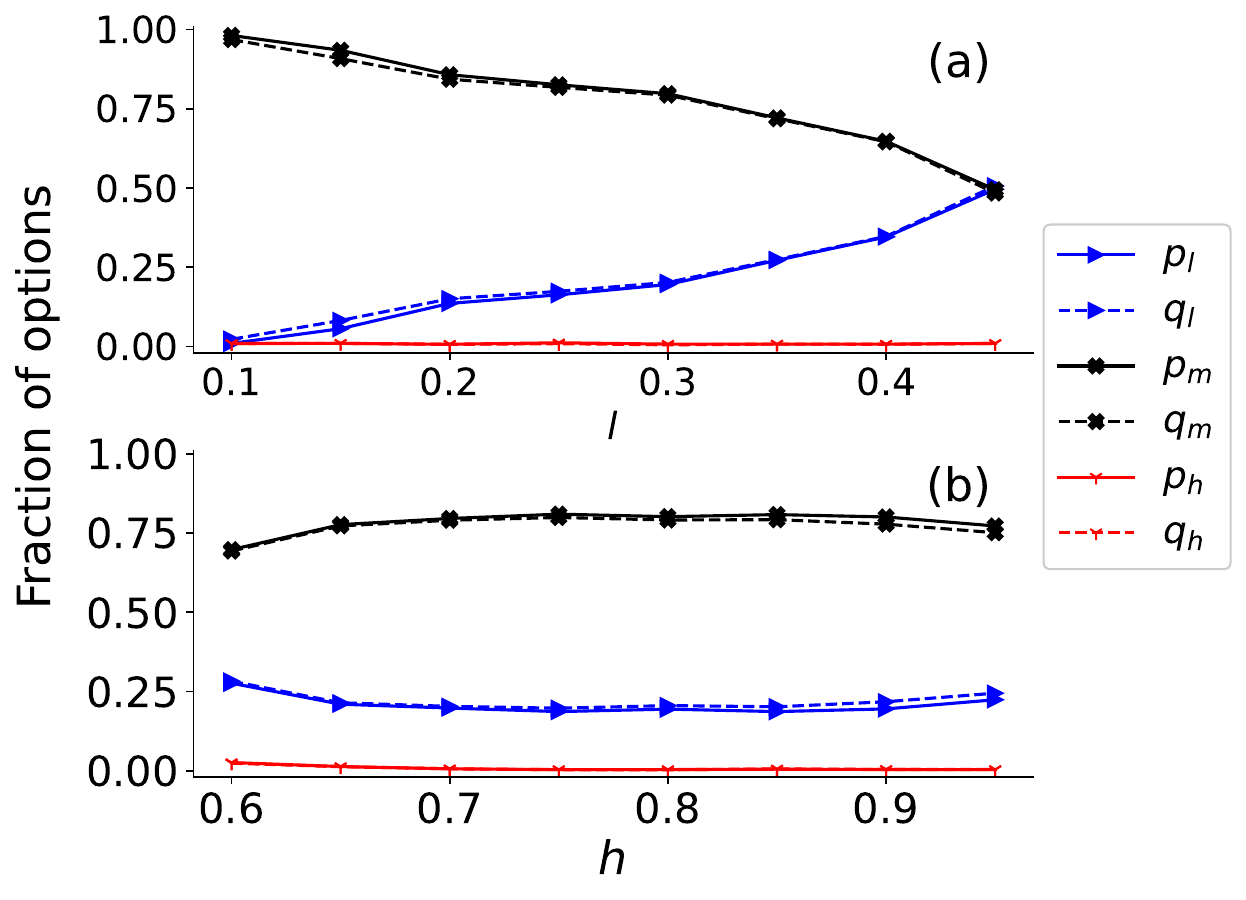} \\ 
\includegraphics[width=0.8\linewidth]{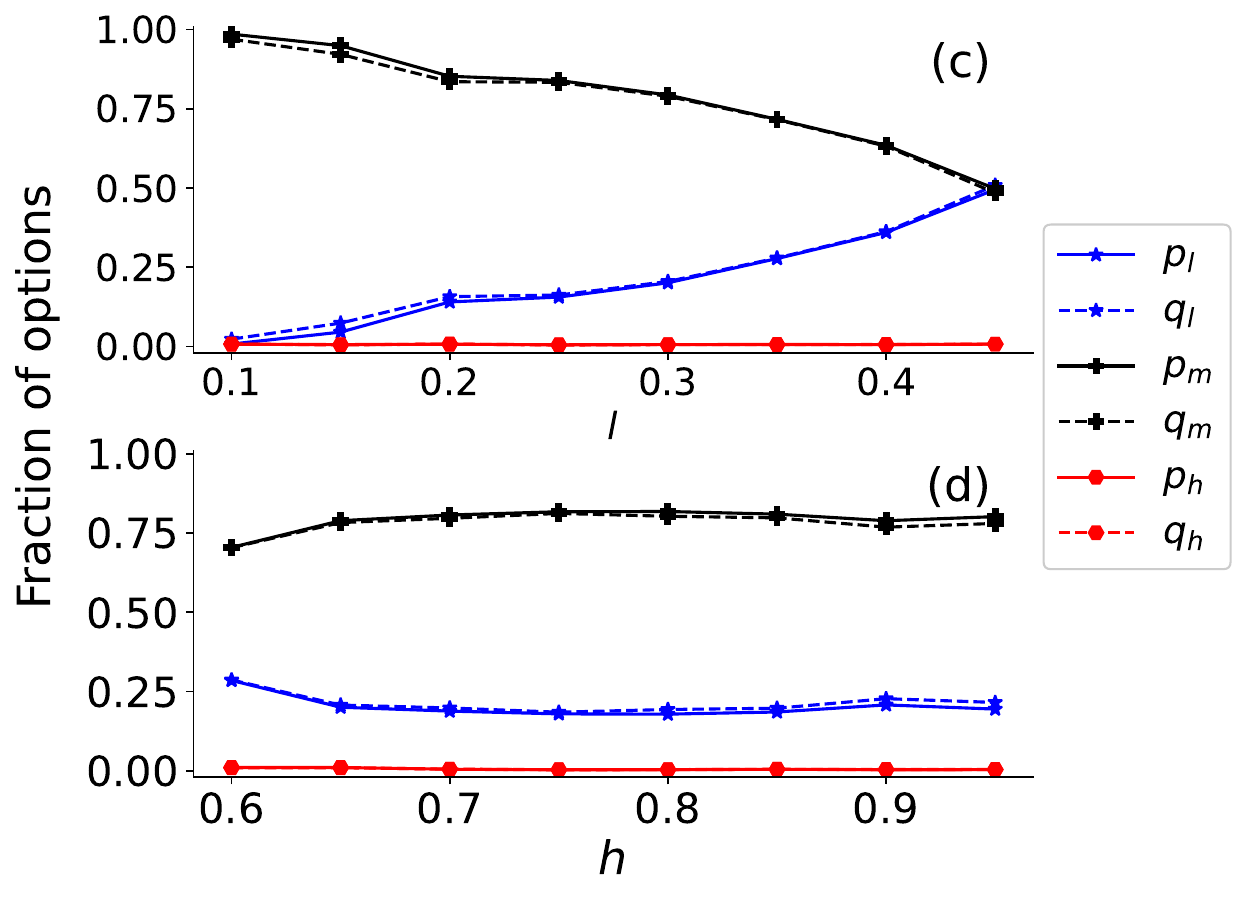}
\caption{\textbf{Different role assignment.} 
The dependencies of fairness on the $l$ and $h$  in the two-player scenario are shown for (a-b) random role and (c-d) fixed role. Other setups are the same as in Fig.~\ref{fig:varylh}.
}
 \label{fig:role_assignment}
\end{figure}

\section{Emergence failure of fairness}\label{sec:failure}

When does fairness fail to emerge within our RL paradigm?
Fig.~\ref{fig:TransitionNetwork} presents the state transition networks for four typical scenarios, only one yielding a decent fairness level [Fig.~\ref{fig:TransitionNetwork}(a)] with the emergence failures within the rest [Fig.~\ref{fig:TransitionNetwork}(b-d)]. Specifically, the state transition network provides the detailed evolution process, where the nodes represent nine states $s_{1,2,...,9}$ with their probabilities being color-coded, and directed edges correspond to the state transitions with the conditional probability as the transition probability $p(s\rightarrow s')=p(s'|s)$ encoded by the line color and width. 
Fig.~\ref{fig:TransitionNetwork}(a) presents the typical evolution paths for $(\alpha, \gamma)=(0.1, 0.9)$, where individuals both appreciate the historical experiences and have the long-term vision. As shown, all nine states finally settled down to either $s_1$ or $s_5$, with dominating paths for the transitions shown, e.g. $p(s_{2,6,8}\rightarrow s_5)$, $p(s_{3,7}\rightarrow s_1)$ are all larger than 0.7. 

However, this is not always the case and the state transition network becomes completely different within the typical failure scenarios, as shown in Fig.~\ref{fig:TransitionNetwork}(b-d). When individuals value historical experiences but are shortsighted [$(\alpha, \gamma)=(0.1, 0.1)$ in Fig.~\ref{fig:TransitionNetwork}(b)], all strategies except for the rational strategies $s_1$ are destabilized. This is because the responder lacks the incentive to adopt a higher acceptance threshold without the guidance of future rewards. %
In the opposite scenario with $(\alpha, \gamma)=(0.9, 0.9)$ as shown in Fig.~\ref{fig:TransitionNetwork}(c), where individuals have high future expectations but rapidly discard historical experiences, the state transition becomes random in the form of scattered paths.  The absence of dominating paths weakens the impact of future guidance, and the emergence of fairness becomes unstable. As a consequence, apart from the fair and rational strategies, the strategy $s_4=(p_m,q_l)$ survives. In particular, $p(s_1\rightarrow s_4)\approx 0.47$ indicates that when the responder adopts strategy $q_l$, the proposer does not necessarily select strategy $p_l$ to maximize the payoff.
Finally, Fig.~\ref{fig:TransitionNetwork}(d) shows the scenario when they both devalue the experience and are short-sighted. By prioritizing immediate payoffs, responders tend to select $q_l$ that favors immediate success so that any offer will lead to a successful deal. However, by disregarding historical experiences, the proposer develops almost no action preference and essentially makes random moves, leading to a fluctuating and diverse pattern of the three strategies $s_1=(p_l,q_l)$, $s_4=(p_m,q_l)$, and $s_7=(p_h,q_l)$.

To further understand the fairness stability of the system, Appendix~\ref{sec:appendixC} provides a theoretical analysis of the stability dependence on learning parameters. This analysis is conducted from the perspective of pathway changes, deriving a predictive boundary that closely aligns with the simulation result. However, this approach fails when applied to the responder.

\section{Robustness}\label{sec:lattice}

Apart from the rotating role assignment, where the two individuals play the two roles in turn, here we also study two other variants and very similar observations are made. Fig.~\ref{fig:role_assignment}(a,b) shows the dependence of the stationary fractions of actions on the values of $l$ and $h$ when roles are randomly assigned. Compared to Fig.~\ref{fig:varylh}, the same observations are made that the value of the low offer $l$ has a significant impact, while the change in overgenerous option $h$ is of marginal impact except at the small end. This is also true for the fixed role setup, where the roles are fixed for the two players, as shown in Fig.~\ref{fig:role_assignment}(c,d).

\begin{figure}[!htbp]
\centering
\includegraphics[width=0.8\linewidth]{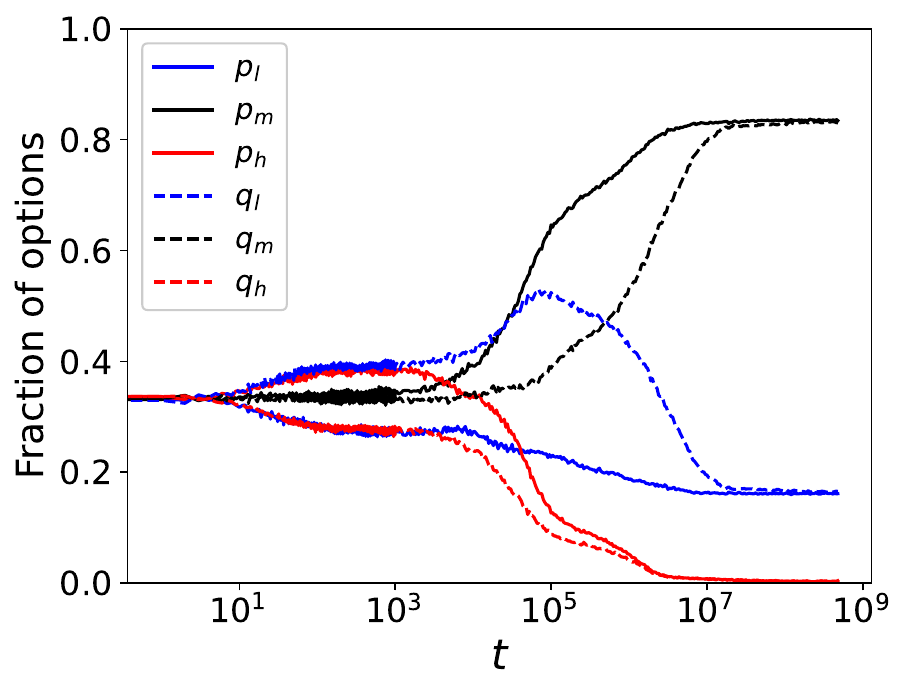}
\caption{\textbf{Time evolution of different actions in 1d latticed population.}  
Similar to the two-player scenario, eventually only the fair strategy $s_5=(p_m, q_m)$ and the rational strategy $s_1=(p_l, q_l)$ survive. Each data is averaged over 100 realizations. 
Other parameters: $\epsilon=0.01$, $\alpha=0.1$, $\gamma=0.9$, $l=0.3$, $h=0.8$, $N=50$, $K=2$.}
\label{fig:lattice}
\end{figure}

These findings are not limited to the two-player scenario studied above; they remain robust when the system is extended to the population level. Fig.~\ref{fig:lattice} shows an example of a one-dimensional latticed population of size $N=50$. In this scenario, each individual has two nearest neighbors ($K=2$), and has separate Q-tables to play the UG with each of them. Eventually, we obtain the time series of the proportions of all six options. We find qualitatively the same result as revealed in the two-player scenario [Fig.\ref{fig:TimeSeries}(a)]. In the long term, the individuals adhere to the fair strategy with a probability exceeding $80\%$, with the rest probability they choose the rational strategy.

\section{Discussion}\label{sec:discussion}

Motivated by the potential of reinforcement learning, we apply it to decoding the emergence of fairness, where exogenous factors are not necessarily involved.
Specifically, we study the ultimatum game with Q-learning, where each player is empowered with Q-tables to guide their decision-making among three typical options. Surprisingly, in the two-player scenario, we find that fairness emerges significantly when individuals appreciate both historical experiences and future rewards. While experiences facilitate individuals to draw lessons from trial and error, the guidance from future rewards prompts responders to shift from the low rational expected offer to the higher fair offer. This then forces the proposer to propose a higher offer to reach a successful deal.

Mechanically, the emergence of fairness consists of two phases. Strategies yielding failed deals are removed from the population in the first phase, and in the second phase the remaining strategies branch into the fair or rational strategies in the end. 
Notably, these findings qualitatively align with existing behavioral experiments on the UG~\cite{Guth1982An, Thaler1988Anomalies,Bolton1995Anonymity, Roth1995The,Henrich2000Does,Oosterbeek2004Cultural,Chuah2007Do,Yamagishi2012Rejection, Guth2014More}. They reveal that proposals exceeding $45\%$ of the total amount are generally accepted, and near-fair strategies are preferred rather than high or low offers. Individuals tend to reject the offer when faced with low proposals.

Crucially, unlike most previous studies grounded in the social learning paradigm, we do not resort to any exogenous factors for explanation. Our results in the RL framework show that endogenous motivation in humans --  maximizing the accumulated rewards -- is sufficient to trigger fair acts. It is also important to point out that the limitation of our study is the discretized setups for both action and state spaces, as they are supposed to be continuous. This is a model simplification for the convenience of investigation and is supposed to be conviently extended to the continuous version by using deep reinforcement learning~\cite{franccois2018introduction}. 

Together with previous studies, our work suggests that the RL framework may offer a unifying paradigm for understanding cooperation~\cite{Zhang2020understanding,Song2022reinforcement,Ding2023emergence}, trust~\cite{Zheng2024Decoding}, and resource allocation~\cite{Andrecut2001q, Zhang2019reinforcement}, and provide new insights into many puzzles in human behaviors.
We acknowledge, however, whether reinforcement learning can serve as a reliable paradigm for interpreting the complexities of human behaviors~\cite{Capraro2018grand, Capraro2021mathematical} requires further validation through behavioral experiments.

\begin{figure}[!htbp]
\centering
    \includegraphics[width=0.9\linewidth]{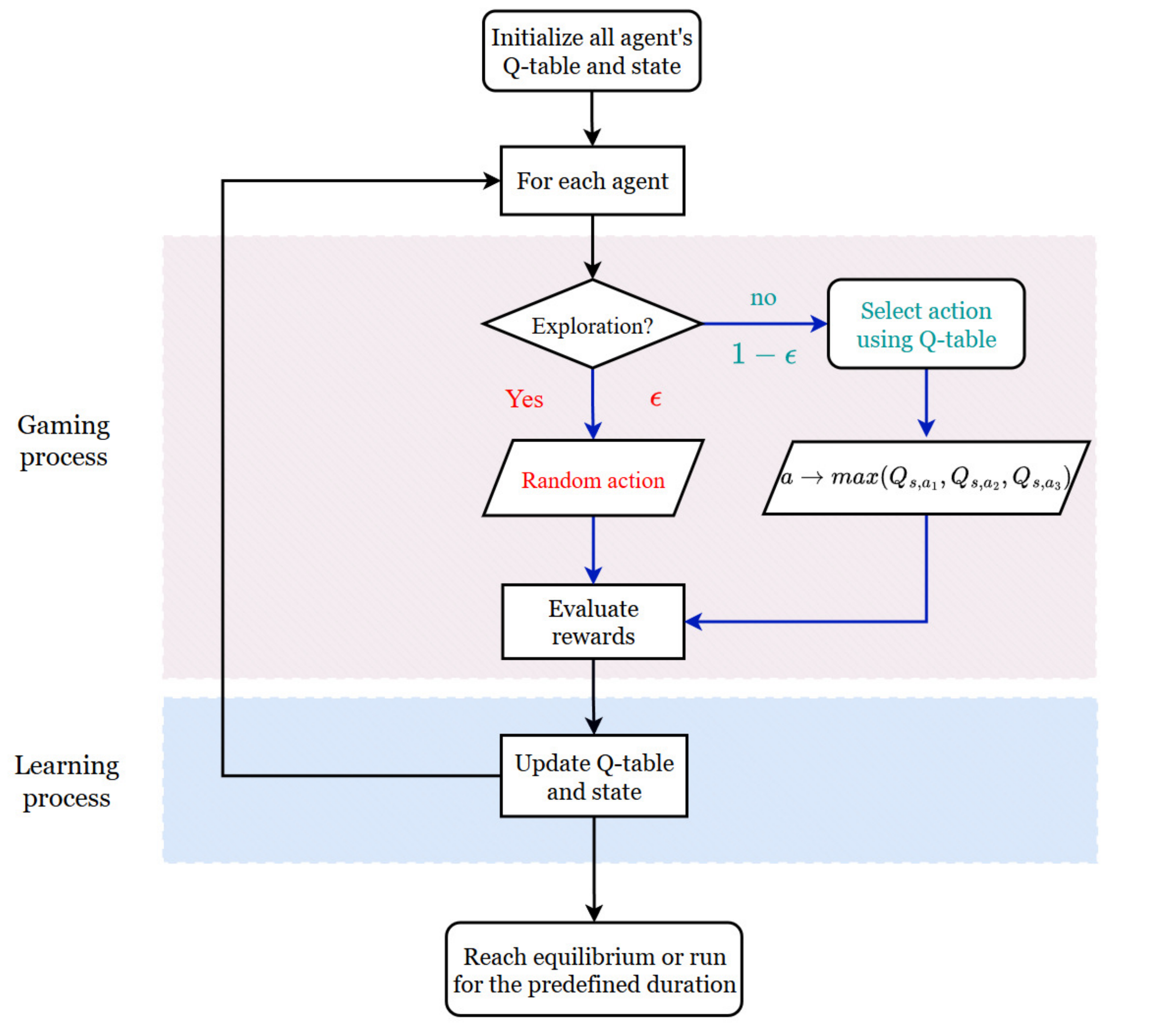}
\caption{Flowchart for the evolution of fairness.}
\label{fig:FlowChart}
\end{figure}

\begin{figure*}[!htbp]
\centering
    \includegraphics[width=0.8\linewidth]{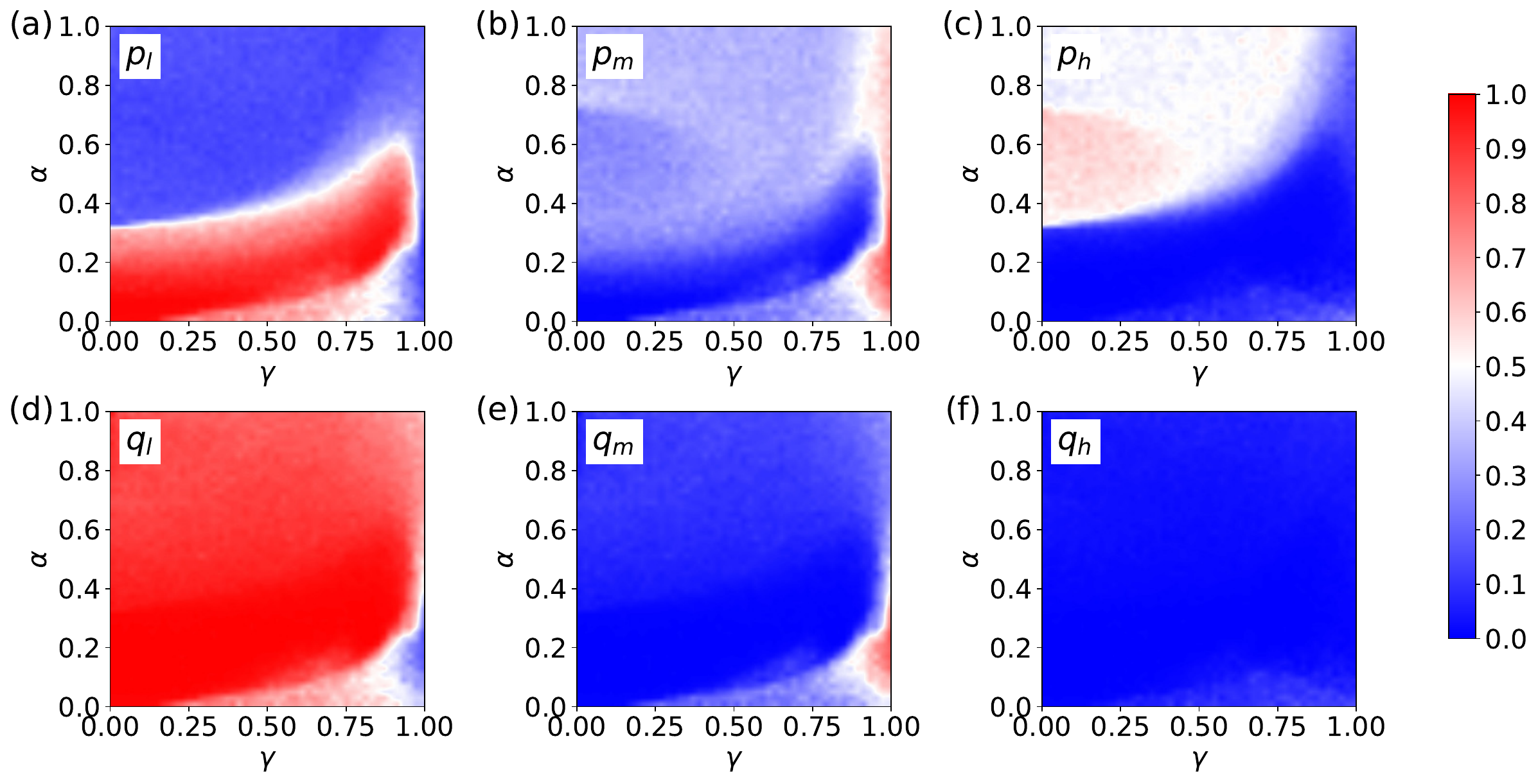}
\caption{The color-coded stationary fractions of the six options in the learning parameter domain ($\gamma,\alpha$). The setup is the same as Fig.~\ref{fig:phasediagram_0.30.50.8}, except for a different parameter combination $l=0.4$ and $h=0.6$ for the game.}
\label{fig:phasediagram_0.40.50.6}
\end{figure*}

\section*{Code availability}
The code for generating key results in this study is available at \href{https://github.com/chenli-lab/RL-Fairness}{https://github.com/chenli-lab/RL-Fairness}.

\section*{Acknowledgments}
This work is supported by the National Natural Science Foundation of China (Grants Nos. 12075144, 12165014), the Fundamental Research Funds for the Central Universities
(Grant No. GK202401002), and the Key Research and Development Program of Ningxia Province in China (Grant No. 2021BEB04032).

\appendix

\section{The evolution protocol}\label{sec:appendixA}

Fig.~\ref{fig:FlowChart} illustrates the evolutionary flowchart of the UG within the reinforcement learning framework. The evolution consists of two processes: the gaming and the learning processes. The details are as follows:
\begin{enumerate}
    \item Initialization: at the beginning, all Q-values are randomly sampled uniformly $Q_{s,a}\in(0,1)$, and the initial states are also selected randomly $s\in\mathbb{S}$.
    \item Gaming process: this includes action selection and reward calculation:
    \begin{itemize}
        \item Action selection: if exploration is chosen, the agent selects an action randomly from the action set with respect to its role; otherwise, the agent selects the action with the highest Q-value in the current state, following the guidance of its Q-table.
        \item Reward calculation: with new actions, and their rewards are then calculated based on Eq.~(\ref{eq:payoff_UG}). 
    \end{itemize}
\item Learning process: each player updates its Q-table based on the rewards obtained, following the Bellman equation described in Eq.~(\ref{eq:Qlearning}) in the main text.
\end{enumerate}

Repeat steps 2 and 3 until the system reaches equilibrium or the predetermined time duration is reached.

\section{Results for game parameters ($l$, $h$) = (0.4, 0.6)}\label{sec:appendixB}

Different from the above study with $(l,m,h)=(0.3,0.5,0.8)$, here we provide the stationary fractions of six strategic options for another set of game parameters $(l,m,h)=(0.4,0.5,0.6)$, shown in Fig.~\ref{fig:phasediagram_0.40.50.6}. 

In Fig.~\ref{fig:phasediagram_0.40.50.6}(b,e), fair acts emerge prominently in the physically meaningful regions, where individuals appreciate historical experiences (a small $\alpha$) and have long-term visions (a large $\gamma$). Conversely, as Fig.~\ref{fig:phasediagram_0.40.50.6}(a,d) shows, when individuals are either forgetful (a large $\alpha$) or short-sighted (a small $\gamma$), rational strategies become more appealing. These findings qualitatively align with the results presented in Fig.~\ref{fig:phasediagram_0.30.50.8}.

However, a closer comparison with Fig.~\ref{fig:phasediagram_0.30.50.8} reveals that the red regions in Fig.~\ref{fig:phasediagram_0.40.50.6}(b,e) shrink. This indicates that the emergence of fairness requires a smaller learning rate $\alpha$ and a larger discount factor $\gamma$. Accordingly, the red regions expand in Fig.~\ref{fig:phasediagram_0.40.50.6}(a,d). This suggests that the proposal with a higher $l$ value is more likely to be accepted. In other words, as $l$ increases, rational strategies are expected to dominate across a broader range of learning parameters.

Additionally, the fraction of $q_h$ remains negligible across the entire learning parameter domain. When proposers emphasize less historical experience, the parameter region corresponding to overgenerous offer $p_h$ is expanded compared to Fig.~\ref{fig:phasediagram_0.30.50.8}(c). This is because the payoff gap between $p_h$ and $p_m$ is narrowed, and thus the option of $p_h$ is often selected.

\begin{figure}[!htbp]
\centering
\includegraphics[width=0.95\linewidth]{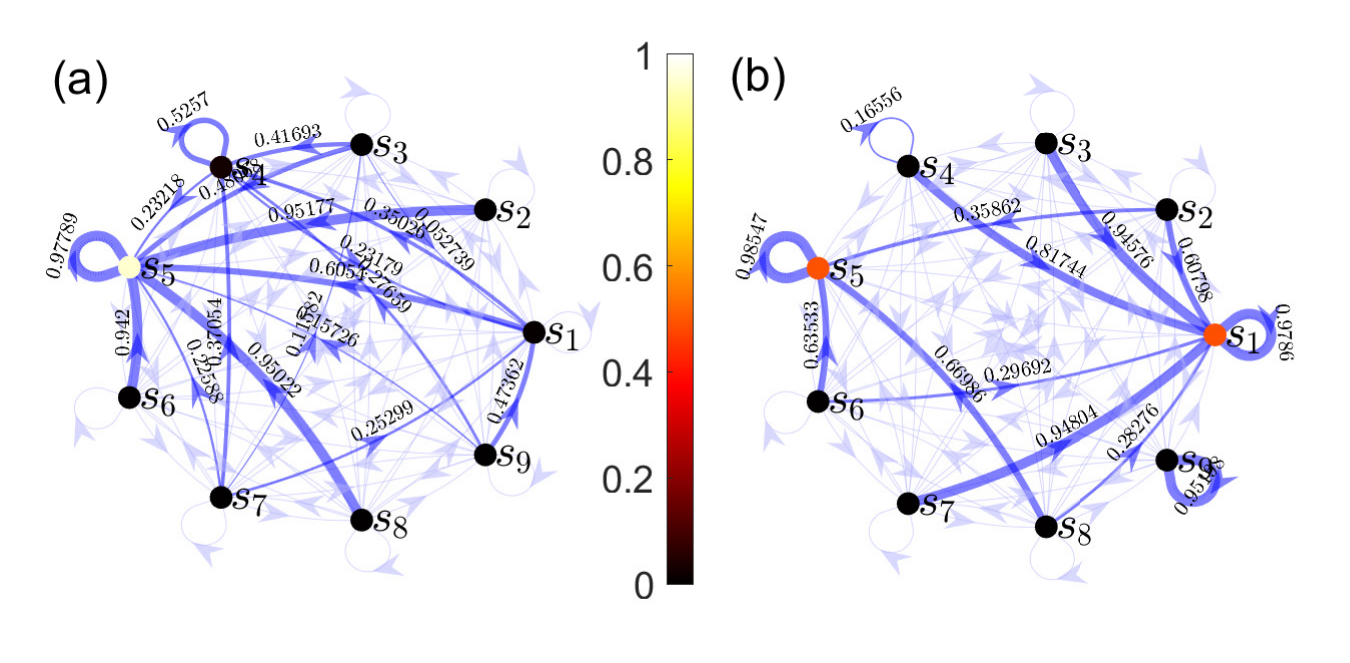}\\
\includegraphics[width=0.95\linewidth]{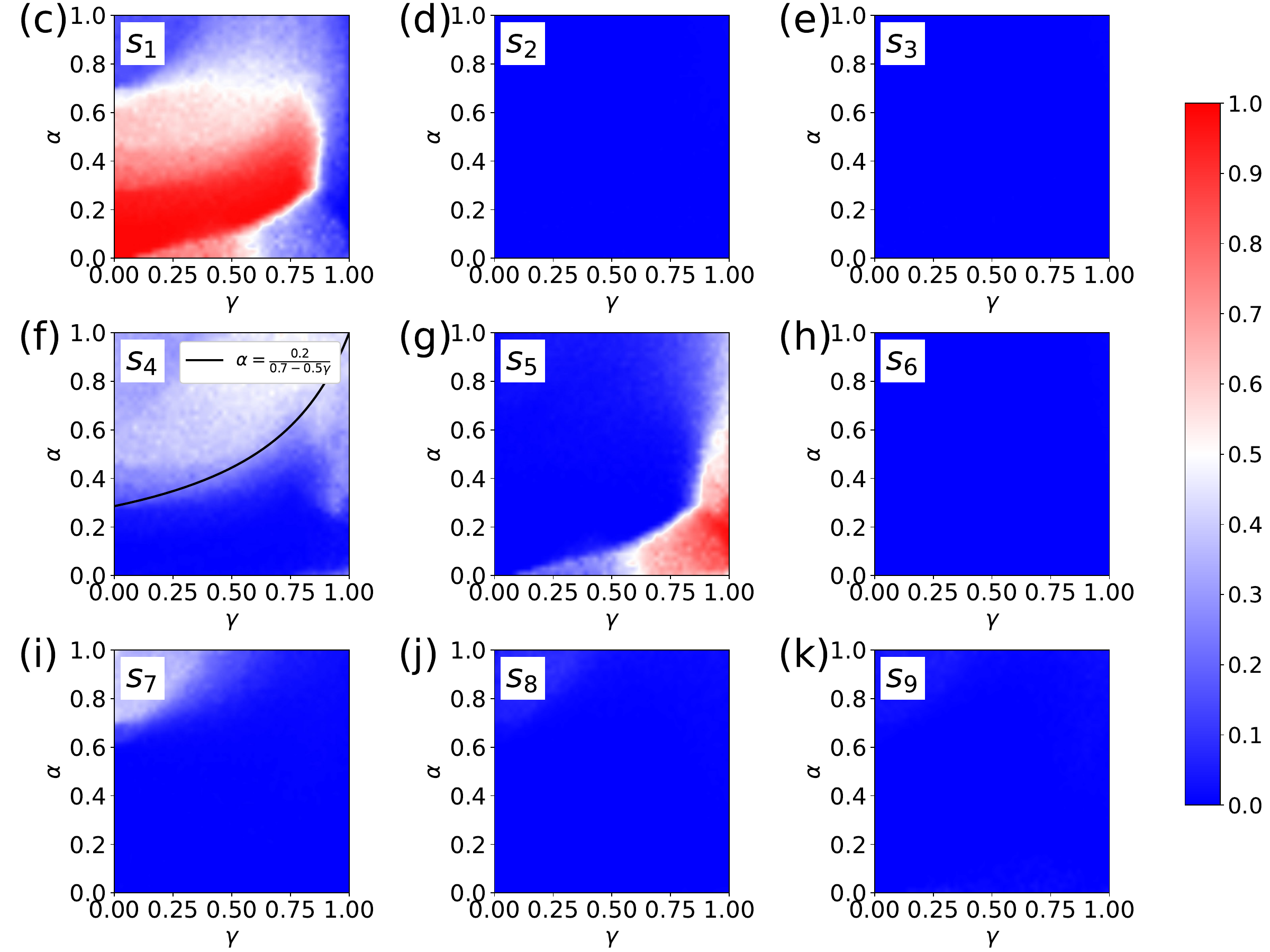}
\caption{The top two panels illustrate state transition networks for the case of $\alpha=0.2$, $\gamma=0.96$ in (a), where fairness emerges stably, as indicated by $f_{p_m,q_m} \rightarrow 1$. The case of $\alpha=0.12$, $\gamma=0.6$ in (b) corresponds to the boundary region of the learning parameters, where fairness and rational strategies are equally likely, i.e., $f_{p_m,q_m} \approx f_{p_l,q_l} \approx0.5$.  Other setups are the same as in Fig.~\ref{fig:TransitionNetwork}.
(c-k) depict the steady-state fractions of nine states, color-coded within the learning parameter domain $(\gamma, \alpha)$, corresponding to states $s_{1, \dots,9}$, respectively. In panel (f), the solid black line indicates the results obtained from the theoretic analysis from Eq.~(\ref{eq:BlackLine}). Each data is averaged over 100 realizations.
Other parameters: $\epsilon=0.1$, $l=0.3$, and $h=0.8$.
}
\label{fig:bounadry_network}
\end{figure}

\section{Boundary analysis}\label{sec:appendixC}

The state transition networks shown in Fig.~\ref{fig:TransitionNetwork} reveal that the evolutionary paths are significantly influenced by the learning parameters, suggesting a critical link between learning parameters and strategy stability. To explore this, we analyze the dependency of fairness stability on learning parameters by comparing the state transition networks of two representative scenarios.
Fig.~\ref{fig:bounadry_network} highlights two scenarios: (a) a region where fairness stably emerges with $f_{p_m,q_m} \rightarrow 1$ [Fig.~\ref{fig:bounadry_network}(a)], and (b) a boundary region where states $s_1=(p_l, q_l)$ and $s_5=(p_m, q_m)$ each account for 50\% of the proportion [Fig.~\ref{fig:bounadry_network}(b)]. 

The comparison reveals that fairness emerges in scenario (a) primarily because, apart from state $s_5$, which is stable, the proportions of all other states approach zero. Additionally, a key pathway, $s_1 \rightarrow s_5$, exists.
In scenario (b), at the boundary, a notable difference is that the self-loop probability of state $s_1$ approaches 1. This indicates that, alongside $s_5$, $s_1$ also becomes stable.

We define $Q_{s, a}^*$ as the stabilized value to which the critical pathway converges upon reaching a stable state. For scenario (a), the critical pathways are as follows: 
\begin{center} 
\begin{tikzpicture}
  \node[circle, draw, minimum size=0.1 cm] (A) at (2, 0) {$s_5$};
  \node[circle, draw, minimum size=0.1 cm] (B) at (0, 0) {$s_1$};

  \draw[->] (B) -- (A) node[midway, above] {};
  
  \draw[->] (A) edge [loop above] ();
\end{tikzpicture}
\end{center}
Consequently, the stabilized Q-values satisfy the following equations: 
\begin{equation}
  \left\{
  \begin{aligned}
Q_{s_5, p_m}^*&\!=\!(1\!-\!\alpha) Q_{s_5, p_m}^*\!+\!\alpha\left(\gamma Q_{s_5, p_m}^*\!+\!\pi_{p_m q_m}\right),\\
Q_{s_1, p_m}^*&\!=\!(1\!-\!\alpha) Q_{s_1, p_m}^*\!+\!\alpha\left(\gamma Q_{s_5, p_m}^*\!+\!\pi_{p_m q_m}\right).
  \end{aligned}
  \right.
  \label{eq:5}
\end{equation}
By computing the fixed point of Eq.~(\ref{eq:5}), we have 
\begin{equation}
  \left\{
  \begin{aligned}
Q_{s_5, p_m}^* &\!=\! \frac{\pi_{p_m q_m}}{1-\gamma}, \\
Q_{s_1, p_m}^* &\!=\! \frac{\gamma \pi_{p_m q_m}}{1-\gamma} + \pi_{p_m q_m}.
  \end{aligned}
  \right.
\end{equation}

Similarly, for scenario (b), the two critical pathways are $s_5 \rightarrow s_5$ and $s_1 \rightarrow s_1$, i.e., 

\begin{center} 
\begin{tikzpicture}
  \node[circle, draw, minimum size=0.1 cm] (A) at (2, 0) {$s_5$};
  \node[circle, draw, minimum size=0.1 cm] (B) at (0, 0) {$s_1$};
  
  \draw[->] (B) edge [loop above] ();
  
  \draw[->] (A) edge [loop above] ();
\end{tikzpicture}
\end{center}
The stabilized Q-values satisfy: 
\begin{equation}
  \left\{
  \begin{aligned}
Q_{s_1, p_l}^*&\!=\!(1\!-\!\alpha) Q_{s_1, p_l}^*\!+\!\alpha\left(\gamma Q_{s_1, p_l}^*\!+\!\pi_{p_l q_l}\right), \\
Q_{s_5, p_m}^*&\!=\!(1\!-\!\alpha) Q_{s_5, p_m}^*\!+\!\alpha\left(\gamma Q_{s_5, p_m}^*\!+\!\pi_{p_m q_m}\right).
  \end{aligned}
  \right.
  \label{eq:7}
\end{equation}
from which we obtain the fixed point as
\begin{equation}
Q_{s_1, p_l}^* = \frac{\pi_{p_l q_l}}{1-\gamma}.
\end{equation}

Therefore, as a proposer, after transitioning into the boundary state \( s_1 \), the decision to change pathways when facing an opponent exploring action \( q_m \) depends on the values of \( Q_{s_1, p_l} \) and \( Q_{s_1, p_m} \), as shown: 
\begin{center} 
\begin{tikzpicture}
  \node[circle, draw, minimum size=0.1 cm] (A) at (0, 0) {$s_1$};
  \node[circle, draw, minimum size=0.1 cm] (B) at (2, 0) {$s_2$};
  \node[circle, draw, minimum size=0.1 cm] (C) at (4, 0) {$s_5$};
  
   \draw[->, dashed] (A) -- (B) node[midway, above] {$\frac{\epsilon}{3}$};
  
  \draw[->] (B) -- (C) node[midway, above] {};
  \draw[->] (C) edge [loop above] ();
\end{tikzpicture}
\end{center}
Specifically, we have:  
\begin{equation}
Q_{s_1,p_l} \!=\! (1\!-\!\alpha) Q_{s_1,p_l}^* \!+\! \alpha (\gamma Q_{s_2,p_m}^* \!+\! \pi_{p_l q_m}) \!=\! Q_{s_1,p_m}^*,
\label{eq:boundary}
\end{equation}
where $s_2=(p_l, q_m)$. In scenario (a), since there exists a pathway $s_2\rightarrow s_5$, we have 
\begin{equation}
  \left\{
  \begin{aligned}
Q_{s_5, p_m}^*&=(1\!-\!\alpha) Q_{s_5, p_m}^*\!+\!\alpha\left(\gamma Q_{s_5, p_m}^*\!+\!\pi_{p_m q_m}\right),\\
Q_{s_2, p_m}^*&=(1\!-\!\alpha) Q_{s_2, p_m}^*\!+\!\alpha\left(\gamma Q_{s_5, p_m}^*\!+\!\pi_{p_m q_m}\right).
  \end{aligned}
  \right.
  \label{eq:9}
\end{equation}
From the Eq.~(\ref{eq:9}), we have $Q_{s_2,p_m}^*=\frac{\gamma \pi_{p_m q_m}}{1-\gamma} + \pi_{p_m q_m}$. Combined with Eq.(\ref{eq:boundary}), the boundary condition for the proposer to choose strategy \( p_m \) or \( p_l \) when facing the responder opts for action \( q_l \) can be derived as:  
\begin{equation}
\alpha =  \frac{\pi_{p_l q_l}-\pi_{p_m q_m}}{\pi_{p_l q_l}-\gamma \pi_{p_m q_m}}.
\label{eq:BlackLine}
\end{equation}
When apply to the game parameters $(l, m, h)=(0.3, 0.5, 0.8)$ used in the main text, the boundary becomes $\alpha=\frac{0.2}{0.7 - 0.5 \gamma}$.

However, when applying this approach to the responder, the scenario changes upon transitioning into the boundary state \( s_1 \). Facing an opponent exploring action \( p_m \), the responder's payoff increases, motivating them to commit to strategy \( q_l \) without any incentive to deviate from this pathway. Consequently, deriving the corresponding boundary condition from the responder's perspective becomes infeasible.

To further validate the results obtained from Eq.(\ref{eq:BlackLine}), subplots (c-k) in Fig.~\ref{fig:bounadry_network} illustrate the color-coded stationary fractions of different states \( s_{1,2,...,9} \) across the parameter domain (\(\gamma,\alpha\)). It can be observed that the strategies primarily appear in subfigures (c) and (f-g), with their corresponding states including options \( p_{l,m} \) and \( q_{l,m} \). Among these, the red regions in subfigures (c,g) closely align with the results of Fig.~\ref{fig:phasediagram_0.30.50.8}(a,e).
Specifically, subplot (d) corresponds to \( s_4 = (p_m, q_l) \). The black solid line represents the theoretical boundary derived from Eq.~(\ref{eq:BlackLine}), which aligns well with the simulation results. However, a noticeable deviation arises as \( \gamma \) approaches 1. We attribute this discrepancy to the slower convergence speed of numerical simulations when \( \gamma \) increases.

\bibliography{fairness}
\end{document}